\documentclass[]{interact}

\usepackage{epstopdf}
\usepackage[caption=false]{subfig}
\usepackage{ae}
\usepackage{graphicx}
\usepackage{epsfig}
\usepackage{amsmath}
\usepackage{amsthm}
\usepackage{amssymb}
\usepackage{subfig}
\usepackage{multirow}
\usepackage{float}
\usepackage{enumitem}
\usepackage{pbox}
\usepackage{hyperref}
\usepackage{endnotes}
\usepackage[table]{xcolor}
\usepackage[numbers,sort&compress]{natbib}
\usepackage[table]{xcolor}
\bibpunct[, ]{[}{]}{,}{n}{,}{,}
\renewcommand\bibfont{\fontsize{10}{12}\selectfont}
\makeatletter
\def\NAT@def@citea{\def\@citea{\NAT@separator}}
\makeatother

\theoremstyle{plain}

\newtheorem{problem}{Problem}
\theoremstyle{definition}

\theoremstyle{remark}
\newtheorem{remark}{Remark}

\let\footnote=\endnote

\begin{document}

\articletype{Full paper manuscript}

\title{Real-time motion planning and decision-making for a group of differential drive robots under connectivity constraints using robust MPC and mixed-integer programming}

\author{
\name{Angelo Caregnato-Neto\textsuperscript{a}\thanks{CONTACT Angelo Caregnato-Neto Email: caregnato.neto@ieee.org}, Marcos O.  A. Maximo\textsuperscript{b} and Rubens J. M. Afonso\textsuperscript{a}}
\affil{\textsuperscript{a}Electronic Engineering Division, Instituto Tecnol\'ogico de Aeron\'autica, São Jos\'e dos Campos, S\~ao Paulo, Brazil; \textsuperscript{b}Autonomous Computational Systems Lab (LAB-SCA), Computer Science Division, Instituto Tecnol\'ogico de Aeron\'autica, São Jos\'e dos Campos, S\~ao Paulo, Brazil}
}

\maketitle

\begin{abstract}
	This work is concerned with the problem of planning trajectories and assigning tasks for a Multi-Agent System (MAS) comprised of differential drive robots. We propose a multirate hierarchical control structure that employs a planner based on robust Model Predictive Control (MPC) with mixed-integer programming (MIP) encoding. The planner computes trajectories and assigns tasks for each element of the group in real-time, while also guaranteeing the communication network of the MAS to be robustly connected at all times. Additionally, we provide a data-based methodology to estimate the disturbances sets required by the robust MPC formulation. The results are demonstrated with experiments in two obstacle-filled scenarios.
\end{abstract}

\begin{keywords}
Model predictive control; trajectory planning; decision-making; connectivity; mixed-integer programming; multi-agent systems
\end{keywords}

\section{Introduction}
The issue of motion planning for groups of autonomous systems, typically denominated Multi-Agent Systems (MAS), has received considerable attention from the academy in recent years \cite{surveyMultiRobot2022,Feola2022}. Such systems leverage their  sheer numbers, flexibility, and robustness to perform tasks that are often unsuitable for a single agent. For example, MAS are suggested to provide enhanced value in coverage applications due to their capacity to spread throughout a huge area and survey more space in less time \cite{drew2021multi}. 

The development of motion planning and decision-making algorithms for MAS is challenging due to the particular issues that arise from the interaction of multiple agents. They must effectively collaborate towards the same global goal, as well as avoid collisions not only with potential obstacles in the environment but also between themselves. Their operation also frequently relies on the maintenance of the communication between the group elements \cite{Afonso2020}, which is typically addressed as a connectivity problem in the graph-theoretical sense \cite{jcae}. 

An attractive method to develop these algorithms is Model Predictive Control (MPC) with Mixed-Integer Programming (MIP) encoding. Motion planners based on MPC-MIP leverage the robustness of the MPC receding horizon strategy and the flexibility provided by a MIP framework, which allows the integration of integer and real-valued decision variables into the optimization problem. Among the many functionalities introduced by MIP models, the capacity of handling the nonconvexity of obstacle-filled environments stands out. Additionally, multiple problems, such as motion planning, decision-making, and connectivity maintenance can be addressed in the same model, allowing for global solutions that take into account all of these aspects jointly. Finally, in spite of its NP-hardness, MIP problems are known to be solved to global optimality in a finite number of iterations. Moreover, through the use of modern solvers and carefully developed models, quality solutions can be computed in real-time \cite{richardsMIPtutorial}.

Early demonstrations of MPC-MIP for MAS motion planning can be traced to \cite{Schouwenaars2001}, where the problem of coordinating multiple agents in obstacle-filled environments is addressed. The joint problem of decision-making, in the form of task allocations, and trajectory planning was addressed for a group of heterogeneous unmanned aerial vehicles (UAVs) in \cite{Bellingham2003}. Further development followed with the introduction of robust MPC \cite{Richards2006} based on the constraint tightening approach. This formulation guarantees that agents under unknown yet bounded disturbances enter a terminal set in finite time. 

A few works also addressed the problem of maintaining connectivity of the communication network of MAS using MIP. In \cite{reinl2007optimal}, a nonlinear optimal control formulation is proposed for the problem of visiting targets in an environment under connectivity constraints. The problem of path planning for a group of UAVs is addressed in \cite{MILPConnConstraints} with a MIP formulation, guaranteeing the connection between a ground station and a target using the agents as relays.  In \cite{Afonso2020} the trajectory planning and task allocation problem for a MAS is solved using a centralized MIP-based planner. The trajectories must guarantee connectivity of the communication network, as well as avoidance of collisions between agents and with obstacles. 

Experimental demonstrations, while not plentiful, are present in the literature. In \cite{Richards2003} an MPC-MIP planner was employed to provide trajectories for a truck in a rendezvous problem considering an obstacle-free environment. A guidance module based on Mixed-Integer Linear Programming (MILP) was proposed in \cite{Schouwenaars2005} and evaluated through coordinated flight tests between an autonomous and a manned aircraft, whereas \cite{Culligan2007} employed a quadrotor to demonstrate an MPC-based trajectory planner with a variable time step formulation. An event-triggered planner is proposed in \cite{anders2017} and used to survey icebergs with a fixed-wing UAV. Fewer works providing experiments with MAS could be found. A trajectory planner module for a group of UAVs is developed using a receding horizon MILP formulation in \cite{how2004flight}. The results are evaluated experimentally in formation flight tests.

\subsection{Contributions}

This work contributes to the subject of MPC-based motion planning by experimentally demonstrating the effectiveness of robust MPC-MIP as a trajectory planning and decision-making technique for a MAS under robust connectivity constraints. To the best of the authors' knowledge, this is the first  experimental demonstration of an MPC-MIP motion planner for MAS under such constraints. Additionally, we provide a data-based methodology for the estimation of the disturbance sets required by the robust MPC formulation.

The demonstrations are made possible through the design of internal controllers that are used to track the trajectories computed by the planner, as well as to cope with the mismatches between the simplified  models used in the MPC formulation and the real dynamics of the robots. The overall architecture, including the planner and the internal controllers, is implemented as a multirate control structure.


\subsection{Notation and Conventions}

 Calligraphic upper case letters, such as $\mathcal{B}$, represent sets. Sets of nonzero positive integers are denoted as  $\mathcal{I}_b = \{1,2,\dots,b \}$. The set difference operation is represented by the symbol $\backslash$. The predicted value of a variable ``$\circ$'', at time step $k+\ell$ is written as $\circ(\ell|k)$. Identity matrices with dimension $n$ are represented by $\mathbf{I}_n$. The term polytope refers only to convex polytopes in this work \cite{verdeArvore}. The time argument $t$ is omitted from continuous-time variables for conciseness. The 2-norm of a vector $\mathbf{v}$ is written as $\Vert \mathbf{v} \Vert_2$. The  Pontryagin difference of two sets $\mathcal{P} \subset \mathbb{R}^n$ and $\mathcal{Q} \subset \mathbb{R}^n$ is defined as \cite{kerriganThesis}:
 \begin{align}
  \mathcal{P} \sim \mathcal{Q} \triangleq \{\omega \in \mathbb{R}^n\ \vert \ \omega + \psi \in \mathcal{P},\ \forall \psi \in \mathcal{Q}\}.
 \end{align}
 Similarly, the Minkowski sum operation is defined as \cite{kerriganThesis}:
  \begin{align}
 	\mathcal{P} \oplus \mathcal{Q} \triangleq \{\delta \in \mathbb{R}^n\ \vert \ \exists \omega \in \mathcal{P}, \phi \in \mathcal{Q}: \delta = \omega + \phi \}.
 \end{align}

\subsection{Outline}
The motion planning problem for a group of differential robots under connectivity constraints is detailed in Section \ref{sec:prob_descrip}. Section \ref{sec:hierarch_ctrl} presents the MPC-MIP formulation and the hierarchical control architecture employed in the experiments. The robust MPC formulation and disturbance set estimation methodology are discussed in Section \ref{section:robust_mpc}. Section \ref{sec:expRes} presents the obtained experimental results. Concluding remarks and future research possibilities are offered in Section \ref{sec:conclusion}.

\section{Problem Description}\label{sec:prob_descrip}

\subsection{Very Small Size Robot}
\begin{figure}[ht!]
	\centering
	\includegraphics[width=0.8\textwidth]{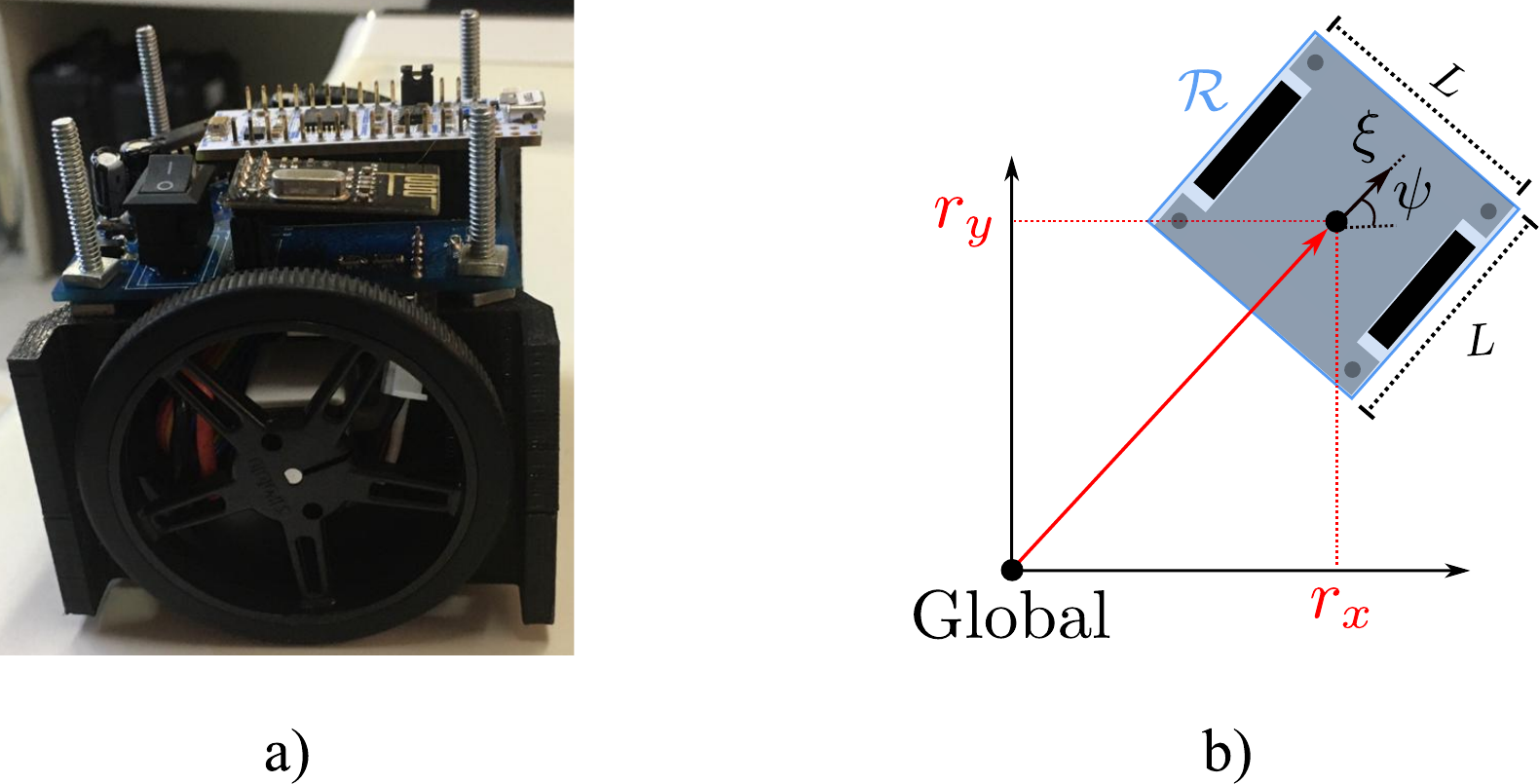}
	\caption{a) Robot from the VSS category and b) corresponding schematic.}
	\label{fig:vss_schem}  
\end{figure}

This work considers a MAS comprised of $n_a$ differential drive robots from the Very Small Size (VSS) category \cite{VSSrules}, as shown in Fig. \ref{fig:vss_schem}a. Their schematic is presented in Fig. \ref{fig:vss_schem}b, where  the polytope $\mathcal{R} \subset \mathbb{R}^2$, which is a square of side $L$, represents their body. Their kinematics is described by the unicycle model
\begin{align}
	&  v_{x,i} = \xi_i \cos(\psi_i), \label{eq:unicycle_1}\\
	&  v_{y,i}  = \xi_i \sin(\psi_i), \label{eq:unicycle_2}\\
	& \dot{\psi_i} = \omega_i,\ \forall i \in \mathcal{I}_{n_a} \label{eq:unicycle_3},
\end{align}
where $v_{x,i}$ and $v_{y,i}$ are the velocities of the $i$-th robot in the $x$ and $y$ axes of a global coordinate system, respectively. The orientation of the same agent is denoted by $\psi_i$, whereas its linear and angular velocities are represented by $\xi_i$ and $\omega_i$, respectively.  The corresponding position vector is denoted by $\mathbf{y}_i = [r_{x,i}, r_{y,i}]^\top$, with $r_{x,i}$ and $r_{y,i}$ being the position coordinates in the $x$ and $y$ axes, respectively.

The dynamic model of the system takes into account the electromechanical dynamics between the input voltage and the torques produced by the motors, as well as the mechanical composition of the robots' bodies.  For a detailed account of the model development the reader is referred to \cite{okuyamaTese}. The ensuing dynamics are represented by the state space equation
\begin{align}
	\boldsymbol{\eta}_i(k+1) = \mathbf{A}_{m} \boldsymbol{\eta}_i(k) + \mathbf{B}_{m}  \mathbf{V}_i(k),\ \forall i \in \mathcal{I}_{n_a},
\end{align}
where $\boldsymbol{\eta}_i = [\dot{\phi}_{r,i},\dot{\phi}_{l,i}]^\top$ is the state vector of the $i$-th robot, with  $\dot{\phi}_{r,i}$  and $\dot{\phi}_{l,i}$ being the corresponding angular velocities of the right and left wheel, respectively. The input vector $\mathbf{V}_i = [V_{r,i},V_{l,i}]^\top$ is comprised of the voltages $V_{r,i}$ and $V_{l,i}$ which are applied to the right  and left motors of the same robot, respectively. Considering a sampling period of $4.2$ ms and a discretization carried out using the Zero-Order Hold (ZOH) method, the state and input matrices are written as
\begin{align}
	\mathbf{A}_{m} = \begin{bmatrix}
		0.9574   & 0.0038\\
		0.0038 &   0.9574
	\end{bmatrix},\ \mathbf{B}_{m} = \begin{bmatrix}
		0.4693 &  -0.0414\\
		-0.0414  &  0.4693
	\end{bmatrix}.
\end{align}

The VSS robots were designed to have a very small clearance from the ground, i.e., their body is constantly in contact with the surface of the field. Thus, the effects of static and kinetic friction are added to the dynamic model as the external input $\mathbf{F}_i \in \mathbb{R}^2,\ \forall i \in \mathcal{I}_{n_a}$, using the friction models developed in \cite{okuyama2017}:
\begin{align}
	\boldsymbol{\eta}_i(k+1) = \mathbf{A}_{m} \boldsymbol{\eta}_i(k) + \mathbf{B}_{m}  (\mathbf{V}_i(k)+ \mathbf{F}_i(k)),\ \forall i \in \mathcal{I}_{n_a}. \label{eq:motor_dyn}
\end{align}

\begin{remark}\label{remark:robotArtesanal}
	The robots were designed and built by students at the Aeronautics Institute of Technology (ITA) during the course \textit{CMC10 -- Design and Manufacturing of Mobile Robots} in 2019. They were constructed with the same general components. Thus, the models employed for the development of the control systems are equal. However, each robot was partially designed and assembled by teams of students with different degrees of experience, resulting in heterogeneous robots with distinctions that are challenging to model. The basic elements of robot's design were developed by the robotics team from ITA, ITAndroids \cite{VSSteamreport}.
\end{remark} 

The actuators of the robots are two brushed motors that operate with voltages within the range of -8.4 V to 8.4 V. The angular velocity of their axes is measured by incremental encoders with a resolution of 12 bits. Each robot is also equipped with a microcontroller that handles the low-level speed control of the motors. A simplex communication system connects the central computer and the group using a radio module. Latency issues are considered negligible. Table \ref{tab:vss_components} presents a detailed account of the components used in the robots.

\begin{table}[!ht]
		\tbl{Main components of each VSS robot.}{
		\begin{tabular}{ccc}\toprule
			Component            &  Specification & Quantity  \\ \midrule
			Motor                & Pololu Micro 50:1  & 2         \\ 
			Radio module         & nRF24L01 & 1               \\ 
			Microcontroller      & STM32F303K8 & 1                 \\ 
			Motor driver         & SN754410 & 1 \\
			Encoder              &  Pololu Encoder Kit Micro & 2               \\ \bottomrule
		\end{tabular}}
	\label{tab:vss_components}
\end{table}

\subsection{Environment}
\begin{figure}[ht!]
	\centering
	\includegraphics[width=0.5\textwidth]{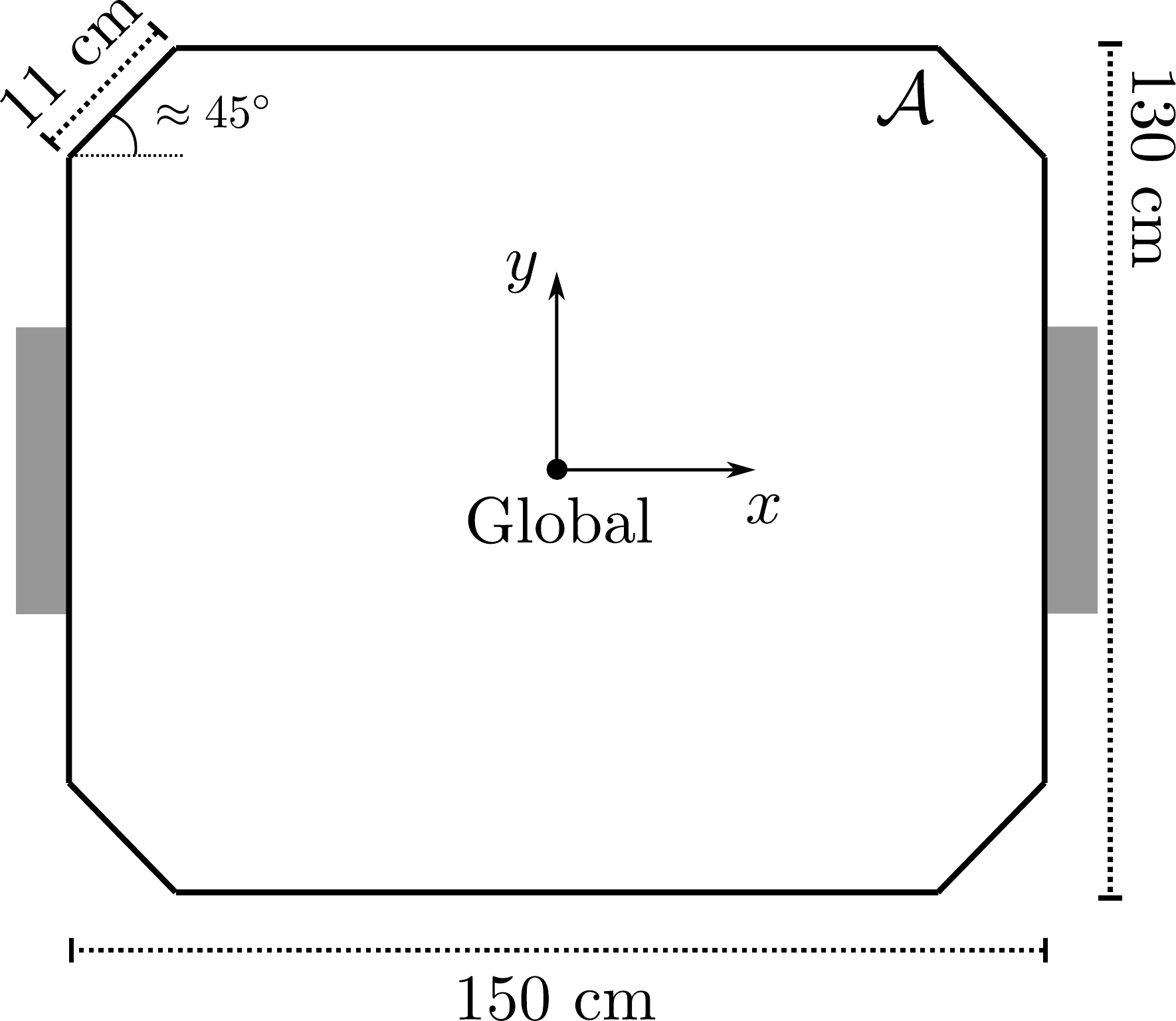}
	\caption{Schematic of the field used as the operational region of the agents. Gray rectangles (soccer goals) are disregarded during the experiments.}
	\label{fig:soccerField_schem}     
\end{figure}
The operational region of the MAS is a field usually employed in the soccer competitions of the VSS league. A schematic with its dimension is presented in Fig. \ref{fig:soccerField_schem}.  The regions representing the soccer goals, depicted in gray, are disregarded as they serve no purpose in the experiments. Thus, the polytope representing the operational region $\mathcal{A}$ corresponds only to the octagon in Fig. \ref{fig:soccerField_schem}. The origin of the global coordinate system is located at the geometric center of the field.

This region is filled with $n_o \in \mathbb{N}$ static obstacles, represented by the polytopes $\mathcal{O}_g \subset \mathbb{R}^2$, $\forall g \in \mathcal{I}_{n_o}$. The space occupied by robot $j$ at time step $k$ is written as:
\begin{align}
	\mathcal{R}_j(k) = \mathbf{y}_j(k) \oplus \mathcal{R}_j,\  \forall k \geq 0.
\end{align}
 The free operational region of the $i$-th robot at time step $k$ is  written as the original operational region without the areas occupied by the obstacles and the remaining robots:
\begin{align}
	\bar{\mathcal{A}}_{i}(k) \triangleq \mathcal{A} \setminus \left \{  \bigcup_{g=1}^{n_o} \mathcal{O}_g \cup \bigcup_{\substack{j\in \mathcal{I}_{n_a}\\ j\neq i}} \mathcal{R}_j(k) \right \},\ i \in \mathcal{I}_{n_a}. \label{eq:free_operation_region}
\end{align}

Within the region $\mathcal{A}$, there are also  $n_t \in \mathbb{N}$ targets represented by the polytopes $\mathcal{T}_{e} \in \mathbb{R}^2$, $\forall e \in \mathcal{I}_{n_t}$.  The mission is  completed when the last target $\mathcal{T}_{n_t}$ is visited by any robot; the remaining targets are optional and must be visited only if advantageous. The objective of the MAS is to complete the mission while optimizing a compromise between the rewards collected by visiting optional targets, the duration of the mission, and the overall control effort.

\subsection{Connectivity}
The time-indexed graphs $\mathcal{G}(k) = \left \{\mathcal{V}  ,\mathcal{E}(k) \right \},\ \forall k\geq 0$, represent the communication network of the MAS, where $\mathcal{V} = \left \{\nu_1,\dots, \nu_{n_a} \right \}$ is the set of vertices and $\mathcal{E}(k) =\left \{(\nu_i,\nu_j)\ |\ \nu_i,\nu_j \in \mathcal{V} \right \}$ is the set of edges. The robots and the connections between them are represented by the vertices and edges of the graph, respectively. The connectivity and robust connectivity of the graph $\mathcal{G}(k)$ imply the same properties for the communication network of the MAS at time step $k$ \cite{jcae}.

 This work considers proximity as the requirement for connections between robots. Thus, a pair of agents are able to communicate if they are inside the connectivity regions of each other. These regions are represented by the polytope $\mathcal{C} \subset \mathbb{R}^2$ and assumed equal for all agents without loss of generality. During the mission, the connectivity regions move since they are centered at the position of the robots. Therefore, they are written as:
\begin{align}
\mathcal{C}_i(k) = \mathbf{y}_i(k) \oplus \mathcal{C},\  \forall i \in \mathcal{I}_{n_a},\ \forall k\geq 0.
\end{align}

 In this context, the main problem addressed by this work is enunciated.
 
 \begin{problem}
 	Develop a trajectory planning and decision-making algorithm for a MAS comprised of the VSS robots such that:
 	\begin{enumerate}[label = {\alph*})]
 		\item the agents collaborate to accomplish the mission while optimizing time expenditure, control effort, and collection of optional rewards;
 		\item state and input bounds are observed;
 		\item collisions with obstacles and between robots are prevented;
 		\item the communication network remains robustly connected.
 	\end{enumerate} 
 \end{problem}
\section{Control Architecture}\label{sec:hierarch_ctrl} 

\begin{figure}[ht!]
	\centering
	\includegraphics[width=1.\textwidth]{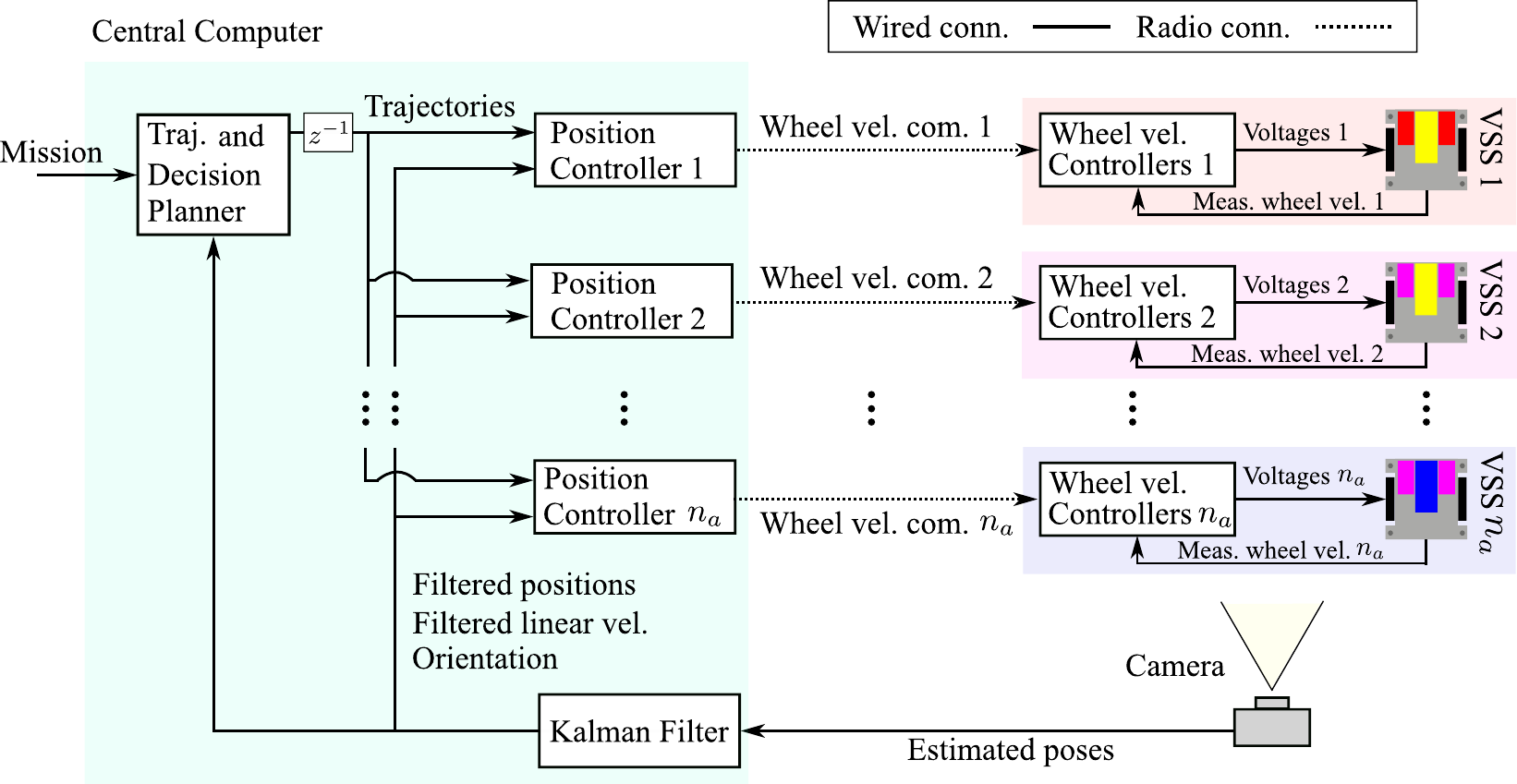}
	\caption{Overview of the proposed control architecture and experiment setup.}
	\label{fig:ctrl_arch_1}     
\end{figure}

The hierarchical control architecture presented in Fig. \ref{fig:ctrl_arch_1}, which employs an MPC-based motion planner, is proposed as a solution to Problem 1. 
The system requires the input of a mission, which consists of information about the operational region, agents' initial positions, and targets. These parameters are fed to the planner that is responsible for outputting trajectories already imbued with the required coordination and decisions. The trajectories are represented by the acceleration, velocity, and position commands of each agent. The system receives the feedback of the measured positions and velocities of each robot and replans their trajectories in real-time. The frequency in which the trajectories are updated is selected as 1 Hz and the resulting sampling period is $T_{plan} = 1$ s.   
This choice represents a compromise between a sufficiently long period, which allows quality solutions to be computed, and relatively regular updates on the trajectories, leveraging the feedback structure of the MPC. Discrete-time variables evaluated at $k_p \in \mathbb{Z}^+$, correspond to continuous-time variables evaluated at $k_pT_{plan}$. 

Typically, the optimization times required by the MPC planner are relatively high and represent a bottleneck in the proposed control architecture. This issue is overcome through the use of a parallel processing scheme and an MPC formulation with delayed inputs, represented by the block $z^{-1}$ in Fig. \ref{fig:ctrl_arch_1}. This issue is discussed in more detail in the subsequent sections.

An intermediary position controller is responsible for computing the necessary linear and angular velocities that must be achieved by the robots such that they follow the trajectories given by the planning module. These controllers are decentralized, with each one receiving the feedback of the corresponding robot's position, velocity, and orientation. Their outputs are the left and right wheel velocity commands of each robot, which are computed using the control law and inverse kinematic equations \cite{Siciliano}. The sampling frequency chosen for each of these control loops is 60 Hz, which makes maximum use of the measurement capacity of the computer vision algorithms available. The corresponding sampling period is $T_{track} = 16.7$ ms. Similarly to the planner loop, discrete-time variables evaluated at $k_t\in \mathbb{Z}^+$ correspond to continuous-time variables at $k_tT_{track}$.  

The wheel velocity commands are transmitted to the MAS through radio communication. They are then tracked by the proportional-integral controllers that are implemented in the embedded microcontrollers of each robot. The controller design was inherited from the development carried out during the \textit{CMC10} course. The proportional and integral gains were selected as $K_{p,w} = 0.0043$ and $K_{s,w} = 0.1981$, respectively. 

Finally, computer vision algorithms are employed to estimate the position and orientation of the agents using the images from a camera. The linear velocities are estimated using a Kalman filter, based on \cite{kfvss}, which also filters the noise from the position measurements. Filtering of the orientations was deemed unnecessary for the correct operation of the proposed control architecture.

\begin{figure}[ht!]
	\centering
	\includegraphics[width=1.\textwidth]{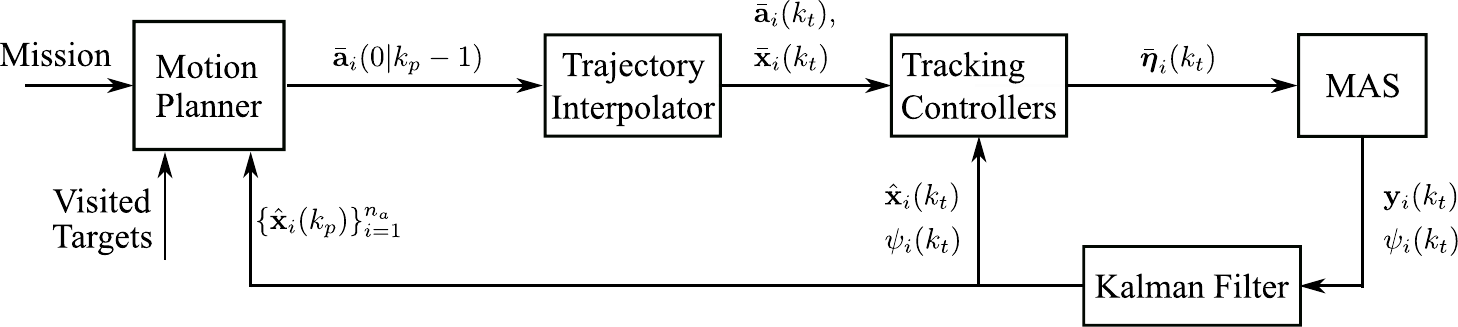}
	\caption{Block diagram of the control architecture, $\forall i \in \mathcal{I}_{n_a}$. Wheel velocity controllers are within the MAS block.}
	\label{fig:ctrl_arch_1_with_vars}     
\end{figure}

The block diagram in Fig. \ref{fig:ctrl_arch_1_with_vars} presents the architecture in detail; in the following section each of its elements is discussed.

\subsection{MPC-MIP Motion Planner} \label{subsec4:delayed_formulation}

As observed in Fig. \ref{fig:ctrl_arch_1_with_vars}, the purpose of the planning module is to provide commands to the trajectory tracking controllers of each agent. To this end, we employ the MPC-MIP  formulation proposed in \cite{jcae}. The following discrete-time state space prediction model is employed:
\begin{align}
	\bar{\mathbf{x}}_i(k_p+1) = \mathbf{A}\bar{\mathbf{x}}_i(k_p) + \mathbf{B}\bar{\mathbf{a}}_i(k_p-1),\ \forall i \in \mathcal{I}_{n_a}, \label{eq:dyn}
\end{align}
where $\bar{\mathbf{x}}_i = [\bar{r}_{x,i},\bar{v}_{x,i},\bar{r}_{y,i},\bar{v}_{y,i}]^\top$ is the state command vector of the $i$-th agent, with $\bar{r}_{x,i}$, $\bar{v}_{x,i}$, $\bar{r}_{y,i}$, $\bar{v}_{y,i}$ being the corresponding position and velocity commands in the $x$ and $y$ axes, respectively. The acceleration command vector is denoted by $\bar{\mathbf{a}}_i = [\bar{a}_{x,i}, \bar{a}_{y,i}]^\top$ with $\bar{a}_{x,i}$ and $\bar{a}_{y,i}$ being the acceleration commands of the $i$-th agent in the $x$ and $y$ axes, respectively. Notice that, in spite of their heterogeneity, all robots are modeled using the same dynamics with state and input matrices:
\begin{align}
	\mathbf{A} = \begin{bmatrix}
		1 & 1 & 0 & 0\\
		0 & 1 & 0 & 0\\
		0 & 0 & 1 & 1\\
		0 & 0 & 0 & 1
	\end{bmatrix},\ \mathbf{B} = \begin{bmatrix}
		0.5 & 0\\
		1 & 0\\
		0 & 0.5\\
		0 & 1
	\end{bmatrix}.	
\end{align}

As previously stated, the acceleration command input in (\ref{eq:dyn}) is delayed by one time step. This is a widely known MPC method used  to handle bottlenecks caused by time-demanding optimizations  \cite{Maciejowski}. We provide more details on its application within this control structure in Section \ref{subsec4:software}. The ensuing optimization problem is now presented.

\begin{remark}
	As stated in Remark \ref{remark:robotArtesanal}, the mechanical characteristics that make the robots distinct are challenging to model and generally minor. Thus, robustness to such uncertainties is expected to be provided by the feedback structure of the MPC and then enhanced by the constraint tightening approach.
\end{remark}

\begin{subequations}
	\begin{alignat}{2}
		&\text{\textbf{Optimization Problem.}} \nonumber \\
		&\underset{\substack{\bar{\mathbf{x}}_i(\ell|k_p),\bar{\mathbf{a}}_i(\ell|k_p),N(k_p),\\ b^{con}(\ell|k_p),b^{tar}(\ell|k_p)}}{\textrm{minimize}}
		\ N(k_p) + \sum_{\ell=0}^{N(k_p)} \sum_{i=1}^{n_a}  \gamma\Vert \bar{\mathbf{a}}_i(\ell|k_p) \Vert_2^2    -\rho \sum_{\ell=0}^{N(k_p)}\sum_{i=1}^{n_a} \sum_{e=1}^{n_t-1} \ b_{i,e}^{tar}(\ell|k_p)  \label{prob1:cost}
		\\
		&\textrm{subject to, } \forall i \in \mathcal{I}_{n_a},\ \forall e\in \mathcal{I}_{n_t},\ \forall \ell \in \{0,\dots,N(k_p)\}, \nonumber \\
		&\bar{\mathbf{x}}_i(0|k_p) = \hat{\mathbf{x}}_{i}(k_p),\label{prob1:const:recHor} \\
		&\bar{\mathbf{x}}_{i}(\ell+1|k_p) = \mathbf{A}\bar{\mathbf{x}}_i(\ell|k_p) + \mathbf{B} \bar{\mathbf{a}}_{i}(\ell-1|k_p), \label{prob1:const:dynamics} \\
		& \bar{\mathbf{a}}_i(-1|k_p) = \bar{\mathbf{a}}_i(0|k_p-1), \label{prob1:const:delayInput}\\
		& \bar{\mathbf{y}}_i(\ell+1|k_p) = \mathbf{C}_i\bar{\mathbf{x}}_i(\ell+1|k_p),\label{prob1:const:output} \\
		&\bar{\mathbf{x}}_{i}(\ell+1|k_p) \in \mathcal{X}_i, \label{prob1:const:state}\\
		&\bar{\mathbf{a}}_{i}(\ell|k_p) \in \mathcal{U}_i, \label{prob1:const:input}\\
		& \mathbf{y}_{i}(\ell+1|k_p) \in \bar{\mathcal{A}}_i(\ell+1|k_p), \label{prob1:const:freeSpace}\\
		& b_{i,e}^{tar}(\ell+1|k_p) \implies \bar{\mathbf{y}}_i(\ell+1 |k_p) \in \mathcal{T}_e, \label{prob1:const:target} \\
		& \sum_{\ell=0}^{N(k_p)}\sum_{i=1}^{n_a} b_{i,e}^{tar}(\ell+1|k_p) \leq 1, \label{prob1:const:target2}\\
		& \exists i\ \vert \ \bar{\mathbf{y}}_{i}(N(k_p)+1|k_p) \in \mathcal{T}_{n_t}, \label{prob1:const:terminal}\\
		&b^{con}_{i,j}(\ell|k_p)  \implies \bar{\mathbf{y}}_i(\ell|k_p)-\bar{\mathbf{y}}_j(\ell|k_p) \in \mathcal{C},\ 1 \leq j < i, \label{prob1:const:conRegion}\\
		&\text{deg}_i(\ell+1|k_p) = \sum_{\epsilon=1}^{i-1}b^{con}_{\epsilon,i}(\ell+1|k_p) + \sum_{\epsilon=i+1}^{n_a}b^{con}_{i,\epsilon}(\ell|k_p), \label{prob1:const:deg}\\
		&\text{deg}_i(\ell|k_p) + \text{deg}_j(\ell|k_p) \geq n_a-n_ab^{con}_{i,j}(\ell|k_p) +\sum_{n=0}^{\ell-1}b^{hor}(n|k_p),\   1 \leq j < i \label{prob1:const:sumVertexDeg}.
	\end{alignat}
\end{subequations}

The cost (\ref{prob1:cost}) represents a compromise between the number of rewards collected by the MAS by visiting optional targets and time and energy expenditure;   $\gamma \in \mathbb{R}^+$ is the fuel weight, whereas  $\rho \in \mathbb{R}^+$ is the reward \cite{Afonso2020}. 
The prediction horizon $N(k_p) \in \{1,2,\dots,N_{max}\}$ in this problem is an optimization variable associated to the condition for mission completion through constraint (\ref{prob1:const:terminal}) \cite{variableHorizon2002}. The variable $N_{max} \in \mathbb{N}$ represents the maximum horizon allowed. 
The receding horizon, dynamics, and output of the system are encoded by constraints (\ref{prob1:const:recHor}), (\ref{prob1:const:dynamics}), and (\ref{prob1:const:output}), respectively.
 In accordance with the delayed input formulation, constraint (\ref{prob1:const:delayInput}) determines that $\bar{\mathbf{a}}_i(-1|k_p),\ \forall k_p \geq 0$,  is always selected as the first element of the sequence of acceleration commands computed at the last time step, $\bar{a}(0|k_p-1)$. 
The bounds on the states and inputs are enforced by (\ref{prob1:const:state}) and (\ref{prob1:const:input}), respectively. 
Collisions with obstacles and between agents are avoided through constraint (\ref{prob1:const:freeSpace}) by guaranteeing that the planned trajectories are always within the free space defined in (\ref{eq:free_operation_region}) \cite{Schouwenaars2001,cornerCutRichards}.
 Constraint (\ref{prob1:const:target}) encodes the condition for activation of the target binaries, i.e., $b^{tar}_{i,e}=1$ implies that robot $i$ is inside target $e$ \cite{Afonso2020}.
 Constraint (\ref{prob1:const:target2}) guarantees that each reward can be collected only once \cite{Afonso2020}. 
 Constraint (\ref{prob1:const:conRegion}) imposes the proximity condition for connections between agents \cite{Afonso2020}. The degree of the vertices representing the robots is encoded using constraint (\ref{prob1:const:deg}). The 2-connectivity conditions that guarantee the robustness of the communication network are enforced by (\ref{prob1:const:sumVertexDeg}) \cite{jcae}. For details regarding this formulation, we refer the reader to the references associated with each element of the optimization problem.

 The employed MPC-MIP formulation  requires the feedback of the state estimate $\hat{\mathbf{x}}_i(k_p)$, $\forall i \in \mathcal{I}_{n_a}$, and the information about the targets that were already visited by the MAS. These targets  are then removed from the optimization problem through relaxations in the constraints.

\begin{remark}
The presented delayed formulation requires the initialization of $\bar{\mathbf{a}}_i(0 |-1)$. This input could be chosen as zero, resulting in the robots remaining still while the first online optimization problem is solved. However, this choice would result in a singularity in the tracking controllers presented in Subsection \ref{subsec4:tracking_ctrl} due to the ensuing null linear velocity commands. Thus, in this work, $\bar{\mathbf{a}}_i(0 |-1)$ is initialized using the solution of the same optimization problem computed \textit{a priori}.
\end{remark}

\subsection{Trajectory Interpolator}


The MPC-MIP planner provides acceleration commands that are updated at each second. Since the tracking control laws operate with a higher frequency, the corresponding velocity and position commands associated with time steps between the MPC updates are computed using the double integrator dynamics discretized with sampling period $T_{track} \leq T_{plan}$.  

As depicted in Fig. \ref{fig:ctrl_arch_1_with_vars}, the ensuing trajectory interpolator block receives the acceleration commands from the planner module and then provides appropriate position and velocity commands to be tracked. Let $\bar{\mathbf{v}}_i = [\bar{v}_{x,i},\bar{v}_{y,i}]^\top$ and $\bar{\mathbf{r}}_i = [\bar{r}_{x,i},\bar{r}_{y,i}]^\top$ denote the vectors of velocity and position commands of the $i$-th agent, respectively. Define the integer $\lambda \in \mathbb{Z}^+$ as $\lambda \triangleq T_{plan}/T_{track}=60$. Then, the propagation equations used to interpolate the trajectories are, $\forall k_p \geq 0$,  $k_p\lambda \leq k_t \leq (k_p+1)\lambda-1$, $\forall i \in \mathcal{I}_{n_a}$,

%
\begin{align}
	&\bar{\mathbf{a}}_i(k_t) = \bar{\mathbf{a}}_i(0|k_p-1),\\
	& \bar{\mathbf{v}}_i(k_t) = \bar{\mathbf{v}}_i(k_t-1) + T_{track}\bar{\mathbf{a}}_i(k_t-1),\\
	& \bar{\mathbf{r}}_i(k_t) = \bar{\mathbf{r}}_i(k_t-1) + T_{track}\bar{\mathbf{v}}_i(k_t-1) +\dfrac{T_{track}^2}{2}\bar{\mathbf{a}}_i(k_t-1).
\end{align}

\subsection{Tracking Controllers} \label{subsec4:tracking_ctrl}

The purpose of the tracking controllers is to compute the appropriate wheel velocity commands for each robot such that they follow the trajectories provided by the planner module. This is accomplished by first computing the required linear and angular velocities of each robot, which are written as the commands $\bar{\xi}_i$ and $\bar{\omega}_i$, and then determining the corresponding wheel velocity commands $\bar{\boldsymbol{\eta}}_i = [\bar{\phi}_{r,i}, \bar{\phi}_{l,i}]^\top$ using inverse kinematic equations \cite{Siciliano}.

The trajectory tracking controllers are designed using the Dynamic Feedback Linearization (DFL) technique \cite{DELUCA1}. This approach allows the development of a compensator that linearizes the kinematic translation equations of the unicycle model exactly, yielding a corresponding closed-loop represented by double integrator dynamics.

Consider a system with kinematics described by the unicycle model, as in (\ref{eq:unicycle_1}), (\ref{eq:unicycle_2}), and (\ref{eq:unicycle_3}). 
Then, by selecting
\begin{align}
	&\dot{\xi}_i = u_{x,i} \cos(\psi_i) + u_{y,i} \sin(\psi_i), \label{eq:nudot}\\
	&\dot{\psi}_i = \omega_i = \dfrac{ -u_{x,i} \sin(\psi_i) + u_{y,i} \cos(\psi_i)}{\xi_i}, \label{eq:nonlin_rot}
\end{align}
one can rewrite (\ref{eq:unicycle_1}) and (\ref{eq:unicycle_2}) as
\begin{align}
	&	 \dot{v}_{x,i} = a_{x,i} = u_{x,i},\\
	&	 \dot{v}_{y,i} = a_{y,i} = u_{y,i},
\end{align}
 where $u_{x,i}$ and $u_{y,i}$ are inputs of the resulting double integrator dynamics and can be designed accordingly to the specifications of the control problem. 



Issues such as discretization, unmodelled effects, and model mismatches could prevent the exact linearization of the translation dynamics. Therefore, the corresponding discrete-time inputs are computed using a proportional-derivative (PD) trajectory tracking control law with feedforward action which is designed to provide additional robustness and improve the overall tracking performance of the system \cite{DELUCA1}

\begin{align}
	& u_{x,i}(k_t) = \bar{a}_{x,i}(k_t) + K_p \left(\bar{r}_{x,i}(k_t) - \hat{r}_{x,i}(k_t) \right) + K_d\left(\bar{v}_{x,i}(k_t) -\hat{v}_{x,i}(k_t) \right), \label{eq:pd_DFL1}\\
	&u_{y,i}(k_t) = \bar{a}_{y,i}(k_t) + K_p\left( \bar{r}_{y,i}(k_t) - \hat{r}_{y,i}(k_t)\right) + K_d\left( \bar{v}_{y,i}(k_t)-\hat{v}_{y,i}(k_t)\right) ,\  \forall i \in \mathcal{I}_{n_a}. \label{eq:pd_DFL2}
\end{align}


The design of the $K_p$ and $K_d$ gains is carried out considering the linear double integrator translation dynamics. However, one must also observe that they have a direct impact on the nonlinear rotation dynamics depicted in (\ref{eq:nonlin_rot}). Thus, the gains are selected as $K_p = 2$ and $K_d=3$ providing an infinite gain margin and a phase margin of $143^\circ$ to the translation system, while also being relatively small, preventing actuator saturation and undesirable behavior of the rotation dynamics.

Finally, the linear and angular velocity commands are computed using equations (\ref{eq:nudot}) and (\ref{eq:nonlin_rot}):
\begin{align}
	&\dot{\bar{\xi}}_i(k_t) = u_{x,i}(k_t) \cos(\psi(k_t)) + u_{y,i}(k_t) \sin(\psi(k_t)),\label{nu_dot}\\
	&  \bar{\omega}_i(k_t) = \dfrac{ -u_{x,i}(k_t) \sin(\psi(k_t)) + u_{y,i} \cos(\psi(k_t))}{\bar{\xi}(k_t)},\label{eq:omega}
\end{align}
with $\bar{\xi}_i$ being computed by integrating (\ref{nu_dot}) numerically. The wheel velocity commands are then determined directly using inverse kinematics equations \cite{Siciliano}.

%
\subsection{Software Implementation} \label{subsec4:software}

Processing of the proposed control architecture is distributed among the  central computer and the embedded microcontrollers of each robot. The latter is responsible only for computations of the control laws related to the tracking of the wheel velocity commands, which are not the main concern of this work and are relatively straightforward software-wise. Thus, only the elements of the former are now discussed.

\begin{figure}[ht!]
	\centering
	\includegraphics[width=0.7\textwidth]{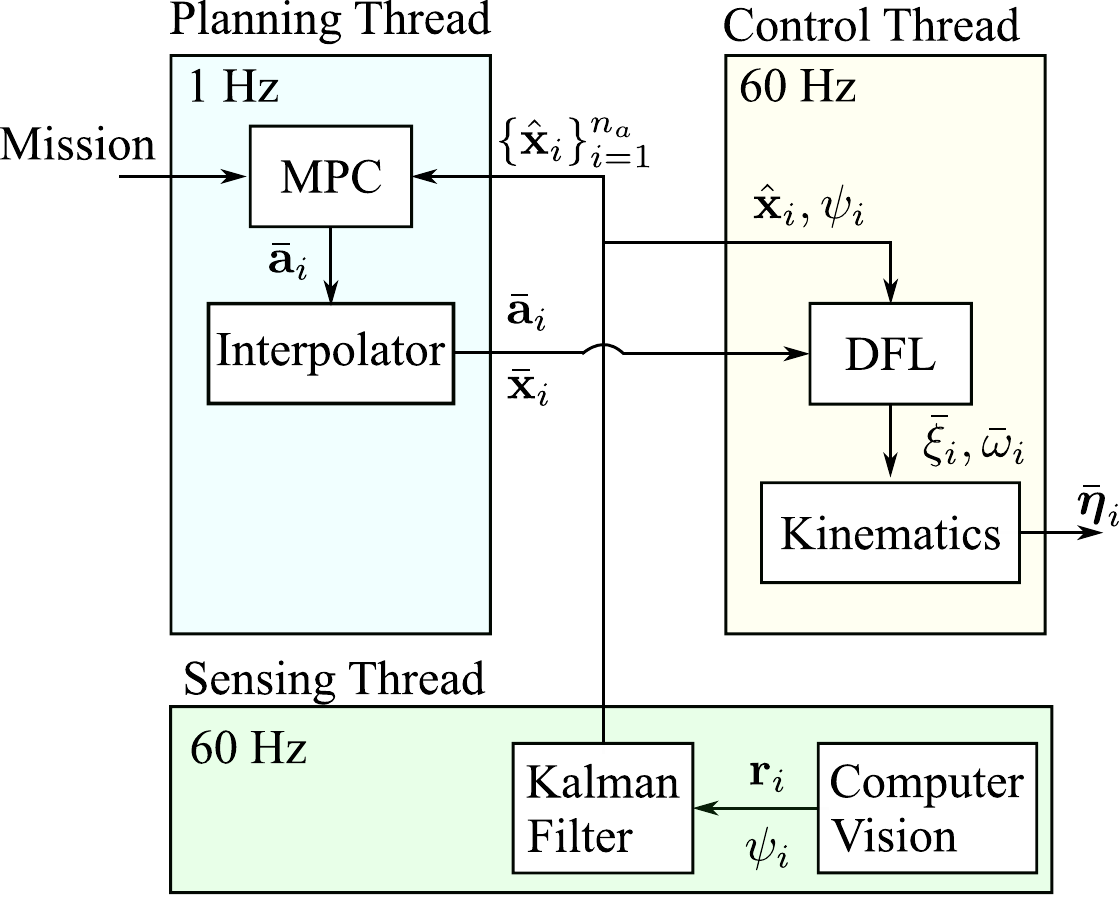}
	\caption{Multirate parallel processing scheme used to implement the MPC-MIP real-time trajectory planning.}
	\label{fig:software_arch}     
\end{figure}

The required computations are divided into three threads that operate in parallel and are responsible for the tasks of planning, control, and sensing. Their interaction is illustrated in Fig. \ref{fig:software_arch}. The planning thread updates the trajectory and decisions of the MAS with a frequency of 1 Hz. Meanwhile, the control thread employs the last solution (determined at the previous planner time step) given by the planning thread to compute the required wheel velocity commands; these commands are updated with a frequency of 60 Hz. Thus, provided that a solution for the MPC-MIP trajectory and decision problem is obtained at each second, the system is able to work in real-time in spite of the bottleneck generated by the relatively high optimization times of the MPC-MIP formulation. Finally, the sensing thread provides estimates of the positions, velocities, and orientations of each robot with a frequency of 60 Hz. All elements in Fig. \ref{fig:software_arch} were implemented for the experiments using C++.

\subsection{Simulation Results}\label{subsec:simVSS}
The proposed control architecture was initially evaluated through a simulation in an environment with $n_a=5$ robots, $n_o=3$ obstacles, and $n_t=3$ targets. The maximum horizon was chosen as $N_{max} = 6$; the control weight and reward were selected as $\gamma=0.2$ and $\rho=3$, respectively. The following bounds were imposed,$ \forall i \in \mathcal{I}_5$,
\begin{align}
	&	-0.75 \  \text{m/s}^2 \leq \bar{a}_{x,i} \leq 0.75\  \text{m/s}^2,\\
	&	-0.75\  \text{m/s}^2 \leq \bar{a}_{y,i} \leq 0.75\  \text{m/s}^2,\\
	&	-0.75\  \text{m/s} \leq \bar{v}_{x,i} \leq 0.75\  \text{m/s},\\
	&	-0.75\  \text{m/s} \leq \bar{v}_{y,i} \leq 0.75\  \text{m/s}. 
\end{align}
The connectivity region of each agent was chosen as a regular octagon with sides of $0.5$ m. Their initial positions are presented in Table \ref{tab:ini_pos}, whereas their initial velocities are zero.  The simulations were carried out using MATLAB$^\text{\textregistered}$. The robots were modeled using the unicycle model and the dynamics presented in (\ref{eq:motor_dyn}). The optimization problem was written with the Yalmip package \cite{Lofberg2004} and solved with Gurobi$^\text{\textregistered}$ \cite{gurobi}. We provide the MATLAB$^\text{\textregistered}$ code employed in the simulations as open source\footnote{\url{https://gitlab.com/caregnato_neto_open/mas_mpc_mip_ita}}.
%
 

\begin{table}[!ht]
\tbl{Initial positions of the robots for simulations and experiments.}
		{\begin{tabular}{ccc}
			\toprule
			Robot & $r_x(0) (m)$ & $r_y(0) (m)$  \\ \midrule
			1     & -0.65       & 0.55          \\ 
			2     & -0.65        & 0.35           \\ 
			3     & -0.40        & 0.15     \\ 
			4     & -0.40         & 0.55         \\ 
			5     & -0.40         & 0.35         \\ \bottomrule
		\end{tabular}}
	\label{tab:ini_pos}
\end{table}

\begin{figure}[ht!]
	\centering
	\includegraphics[width=0.75\textwidth]{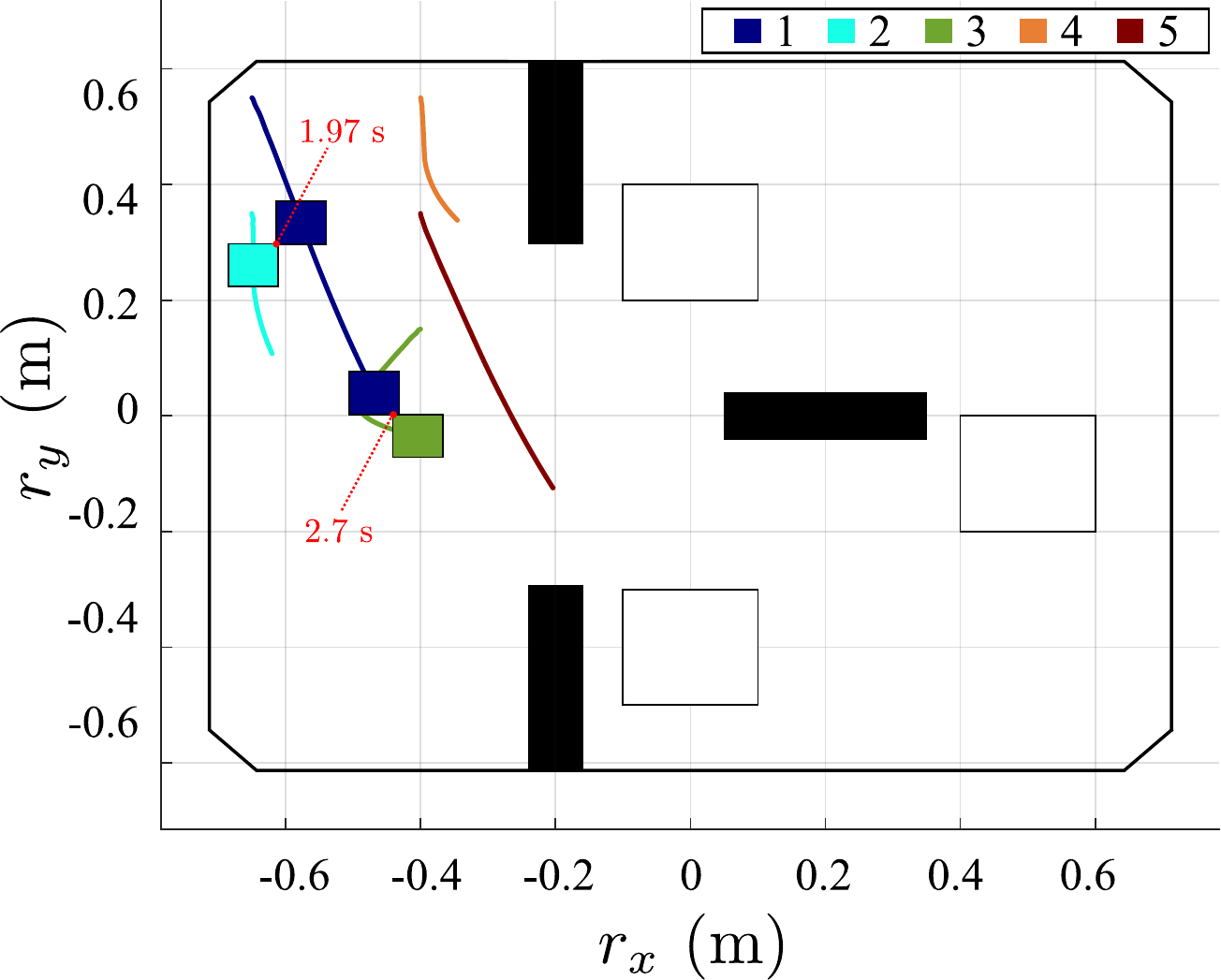}
	\caption{Simulation results of the proposed control architecture in scenario 1.  Collisions between agents 1, 2, and 3 occur during the initial time steps of the maneuver.}
	\label{fig:sim_col}
\end{figure}

The trajectory followed by the robots at the initial time steps is presented in Fig. \ref{fig:sim_col}. As it can be observed, two collisions, between robots 1 and 2 at $t = 1.97$ s and between robots 1 and 3 at $t=2.7$ s occurred. Section \ref{section:robust_mpc} introduces robust MPC as a solution to this issue.

\section{Robust MPC} \label{section:robust_mpc}

\subsection{Constraint Tightening}

The robust MPC formulation is implemented using  the  method introduced by  \cite{CT_orig} and extended in \cite{CT_ext,Richards2006} where the aforementioned unaccounted factors are approximated as a state disturbance and added to the dynamic model
\begin{align}
	&	\bar{\mathbf{x}}_i(\ell+1|k_p)  = \mathbf{A}\bar{\mathbf{x}}_i(\ell|k_p) + \mathbf{B}\bar{\mathbf{a}}_i(\ell-1|k_p)
	+\mathbf{w}_i(k_p),\\
	& \bar{\mathbf{y}}_i(k_p) = \mathbf{C}\bar{\mathbf{x}}_i(k_p),\ \forall i \in \mathcal{I}_{n_a},
\end{align}
where $\mathbf{w}_i = [w_{rx,i}, w_{vx,i}, w_{ry,i}, w_{vy,i}]^\top$ and $\bar{\mathbf{y}}_i = [\bar{r}_{x,i},\bar{r}_{y,i}] ^\top$ are the  disturbance  and position command vectors of the $i$-th agent, respectively. It is assumed that $\mathbf{w}_i(k_p) \in \mathcal{W}_i,\ \forall k_p \geq 0$, with $\mathcal{W}_i \subset \mathbb{R}^{n_x},\ \forall i \in \mathcal{I}_{n_a}$ being the corresponding disturbance polytope. In view of the heterogeneity of the MAS, a distinct polytope $\mathcal{W}_i$  was determined for each agent.

The first step in developing the robust MPC formulation is to rewrite the expression that represents the free operation region (\ref{eq:free_operation_region}) in terms of the tightened polytopes, $\forall i \in \mathcal{I}_{n_a}$, $\forall \ell \in \{0,1,\dots,N(k_p)\}$,

\begin{align}
	\bar{\mathcal{A}}_{i}(\ell|k_p) \triangleq \mathcal{A}_i(\ell|k_p) \setminus \left \{  \bigcup_{g=1}^{n_o} \mathcal{O}_{g,i}(\ell|k_p) \cup \bigcup_{\substack{j\in \mathcal{I}_{n_a}\\ j\neq i}} \mathcal{R}_{ij}(\ell|k_p) \right \}, \label{eq:tight_op_region}
\end{align}
where
\begin{align}
	&\mathcal{A}_i(\ell+1|k_p) = \mathcal{A}_i(\ell|k_p) \sim \mathbf{CL}(\ell)\mathcal{W}_i, \\
	& \mathcal{O}_{g,i}(\ell+1|k_p) = \mathcal{O}_{g,i}(\ell|k_p)  \oplus \mathbf{CL}(\ell)\mathcal{W}_i,\  \forall g \in \mathcal{I}_{n_o},\\
	& \mathcal{R}_{ij}(\ell+1|k_p) = \bar{\mathbf{y}}_j(\ell+1|k_p) \oplus \mathcal{R} \oplus \mathbf{CL}(\ell)\mathcal{W}_{ij},\ 1 < j \leq i,\ \mathcal{R}_{ij} = \mathcal{R}_{ji},
\end{align}
with 
\begin{align*}
	&\mathbf{L}(\ell+1) = (\mathbf{A}-\mathbf{B}\mathbf{K}_{CT})\mathbf{L}(\ell),\ \mathbf{L}(0) = \mathbf{I}_{n_x},\\
	& \mathcal{W}_{ij} = \mathcal{W}_i \oplus \mathcal{W}_j.
\end{align*}

The candidate controller is selected as a static feedback gain $\mathbf{K}_{CT} \in \mathbb{R}^{2 \times 4}$ designed using the Linear Quadratic Regulator (LQR) technique with the cost
\begin{align}
	\sum_{k_p=0}^\infty \left (\bar{\mathbf{x}}(k_p)^\top \boldsymbol{\mu}_{state} \bar{\mathbf{x}}(k_p) + \bar{\mathbf{a}}(k_p)^\top \boldsymbol{\mu}_{input} \bar{\mathbf{a}}(k_p) \right).
\end{align}
The same design was employed for each agent. The weights were chosen as
\begin{align}
	\boldsymbol{\mu}_{state} = \begin{bmatrix}
		100 & 0 & 0 & 0\\
		0 & 0.1 & 0 & 0	\\
		0 & 0 & 100 & 0\\
		0 & 0 & 0 & 0.1
	\end{bmatrix},\  \boldsymbol{\mu}_{input} = \begin{bmatrix}
		0.001 & 0 \\ 
		0 & 0.001
	\end{bmatrix}.
\end{align}

As observed in (\ref{eq:tight_op_region}), the main distinction of the new tightened terrain is that it \textit{shrinks} as the obstacle and collision avoidance regions expand and the bounds on the positions tighten at each time step.

Similarly, the target sets and connectivity regions are also tightened, $\forall i \in \mathcal{I}_{n_a}$, $\forall \ell \in \{0,1,\dots ,N(k_p)\}$,
\begin{align}
	&\mathcal{T}_{e,i}(\ell+1|k_p) = \mathcal{T}_{e,i}(\ell|k_p) \sim  \mathbf{CL}(\ell)\mathcal{W}_i,\ \forall e \in \mathcal{I}_{n_t},\\
	& \mathcal{C}_{ij}(\ell+1|k_p) = \bar{\mathbf{y}}_i(\ell+1|k_p) \oplus \mathcal{C} \sim (\mathbf{CL}(\ell)\mathcal{W}_{ij}),\ 1 < j \leq i,
\end{align}
with $\mathcal{C}_{ij} = \mathcal{C}_{ji}$. Finally, constraints (\ref{prob1:const:target}), (\ref{prob1:const:terminal}), (\ref{prob1:const:conRegion}) are respectively rewritten as
\begin{align}
	& b_{i,e}^{tar}(\ell+1|k_p) \implies \bar{\mathbf{y}}_i(\ell+1 |k_p) \in \mathcal{T}_{e,i}(\ell+1|k_p), \label{const:target_tight}\\
	& \exists i\ \vert \ \bar{\mathbf{y}}_{i}(N(k_p)+1|k_p) \in \mathcal{T}_{n_t,i}(N(k_p)+1|k_p), \label{const:terminal_tight}\\
	&  b^{con}_{i,j}(\ell|k)  \implies \bar{\mathbf{y}}_i(\ell|k)-\bar{\mathbf{y}}_j(\ell|k) \in \mathcal{C}_{ij}(\ell+1|k),\ 1 < j \leq i,
\end{align}
whereas constraint (\ref{prob1:const:freeSpace}) is implemented using the tightened free operation region polytope  (\ref{eq:tight_op_region}). For useful discussions on the use of set theory in constrained optimization problems the reader is referred to \cite{Kolmanovsky1998,Richards2006,Baotic2009}.

\subsection{Disturbance Set Estimation}\label{subsec:distSet_est}
The discussion of the procedure to estimate $\mathcal{W}_i$, $\forall i \in \mathcal{I}_{n_a}$, is presented now for a single agent and the indices $i$ are dropped from the variables for simplicity. The estimates of all disturbance sets can be determined by repeating this procedure individually for each robot.

We start by describing how to compute the expected robot trajectory given a sequence of acceleration commands. The expected closed-loop translation dynamics of the differential robots under the DFL control law is
\begin{align}
	&  a_x = \bar{a}_x + K_p(\bar{r}_x-r_x) + K_d(\bar{v}_x-{v}_x),\\
	&  a_y = \bar{a}_y + K_p(\bar{r}_y-r_y) + K_d(\bar{v}_y-{v}_y), 
\end{align}
which is rewritten in state space form as
\begin{align}
	& \dot{\boldsymbol{\sigma}} = \boldsymbol{\Gamma}_c\boldsymbol{\sigma} + \boldsymbol{\Phi}_c\bar{\mathbf{a}},\\
	& \boldsymbol{\zeta} = \boldsymbol{\Omega}_c \boldsymbol{\sigma},
\end{align}
with $\boldsymbol{\sigma} = [r_x,v_x,r_y,v_y,\bar{r}_x,\bar{v}_x,\bar{r}_y,\bar{v}_y]^\top$, $\bar{\mathbf{a}}=[\bar{a}_x,\bar{a}_y]^\top$ , and  $\boldsymbol{\zeta} = [r_x,v_x,r_y,v_y]^\top$  being state, input, and output  vectors, respectively. The state and input matrices are
\begin{align*}
	\boldsymbol{\Gamma}_c = \begin{bmatrix}
		0 & 1 & 0 & 0 & 0 & 0 & 0 & 0\\
		-K_p & -K_d & 0 & 0 & K_p & K_d & 0 & 0\\
		0 & 0 & 0 & 1 & 0 & 0 & 0 & 0\\
		0 & 0 & -K_p & -K_d & 0 & 0 & K_p & K_d\\
		0 & 0 & 0 & 0 & 0 & 1 & 0 & 0\\
		0 & 0 & 0 & 0 & 0 & 0 & 0 & 0\\
		0 & 0 & 0 & 0 & 0 & 0 & 0 & 1\\
		0 & 0 & 0 & 0 & 0 & 0 & 0 & 0
	\end{bmatrix},\ \boldsymbol{\Phi}_c = \begin{bmatrix}
		0 & 0 \\
		1 & 0 \\
		0 & 0 \\
		0 & 1 \\
		0 & 0 \\
		1 & 0 \\
		0 & 0 \\
		0 & 1 
	\end{bmatrix}.
\end{align*}

For the sake of consistency and simplicity, the system is discretized using the same sampling period $T_{plan}$. Thus, the ensuing disturbance estimates correspond directly to $\mathbf{w}(k_p)$. The equations of the discretized system are
\begin{align}
	& \boldsymbol{\sigma}(k_p+1) = \boldsymbol{\Gamma}\boldsymbol{\sigma}(k_p) + \boldsymbol{\Phi}\bar{\mathbf{a}}(k_p),\label{eq:DLF_cl}\\
	& \boldsymbol{\zeta}(k_p) = \boldsymbol{\Omega} \boldsymbol{\sigma}(k_p).
\end{align}

Let $\boldsymbol{\zeta}_m = [r_{x,m},v_{x,m},r_{y,m},v_{y,m}]^\top$ denote the vector of positions and velocities measured during an experiment  and define $\boldsymbol{\sigma}_m \triangleq [\boldsymbol{\zeta}_m, \bar{r}_x, \bar{v}_x, \bar{r}_y, \bar{v}_y]^\top$. Then, given the corresponding sequence of acceleration commands used in the experiment, one can compute the disturbance estimate as, $\forall k_p \geq 0$,
\begin{align}
	&\hat{\mathbf{w}}(k_p) \triangleq \boldsymbol{\zeta}_m(k_p+1) - \boldsymbol{\Omega}(\boldsymbol{\Gamma}\boldsymbol{\sigma}_{m}(k_p)+\boldsymbol{\Phi}\bar{\mathbf{a}}(k_p)). \label{eq:dist_est}
\end{align}
Thus, the estimate is determined by subtracting the expected state at $k_p+1$ given the measured state $\boldsymbol{\sigma}_{m}(k_p)$, the input $\bar{\mathbf{a}}(k_p)$, and the dynamics (\ref{eq:DLF_cl}), from the measured state at $k_p+1$, $\boldsymbol{\zeta}_m(k_p+1)$.

\begin{remark}
	The one time step ahead propagation of the measured state is employed to determine $\hat{\mathbf{w}}$ as the use of the nominal $\boldsymbol{\sigma}$ would result in an estimate that encompasses the effect of previous disturbances.
\end{remark}


Once a sequence of $n_d \in \mathbb{N}$ disturbance estimates $\{ \hat{\mathbf{w}}(k_p)\}_{k_p=0}^{n_d}$ are determined, one must select an appropriate polytope $\mathcal{W}$ that represents the set containing all potential disturbances that are expected to affect the agent during its operation. There are two properties for $\mathcal{W}$ that should be observed. First, it must contain the origin $\mathbf{w} = [0,0,0,0]^\top$. Second, as it is impractical to explore all potential behaviors of the robots when collecting the data to determine the estimates, $\mathcal{W}$ is selected to be symmetrical w.r.t. the origin. 
Finally, define
\begin{align}
	&	\sigma_{rx} \triangleq 	\max_{k_p \in \{0,1,\dots,n_d\}}  \vert \hat{w}_{rx}(k_p) \vert ,\\
	& \sigma_{vx} \triangleq 	\max_{k_p \in \{0,1,\dots,n_d\}}  \vert \hat{w}_{rv}(k_p) \vert,\\
	&	\sigma_{ry} \triangleq 	\max_{k_p \in \{0,1,\dots,n_d\}}  \vert \hat{w}_{ry}(k_p) \vert,\\
	&	\sigma_{vy} \triangleq 	\max_{k_p \in \{0,1,\dots,n_d\}}  \vert \hat{w}_{vy}(k_p) \vert.
\end{align}
Then, the disturbance polytope is chosen as
\begin{align}
	\mathcal{W} = \{\mathbf{w}\ |\ \mathbf{P}_w \mathbf{w} \leq \mathbf{q}_w \},
\end{align}
with
\begin{align*}
	\mathbf{P}_w = \begin{bmatrix}
		\mathbf{I}_2 &  \mathbf{0}_{2\times 2}\\
		-\mathbf{I}_2 &  \mathbf{0}_{2\times 2}\\
		\mathbf{0}_{2\times 2} &  \mathbf{I}_2\\
		\mathbf{0}_{2\times 2} &  -\mathbf{I}_2
	\end{bmatrix}, 
	\mathbf{q}_w = \begin{bmatrix}
		\sigma_{rx}\\
		\sigma_{vx}\\
		-\sigma_{rx}\\
		-\sigma_{vx}\\
		\sigma_{ry}\\
		\sigma_{vy}\\
		-\sigma_{ry}\\
		-\sigma_{vy}
	\end{bmatrix}.
\end{align*}

\subsection{Data Collection}
In order to collect the data necessary to estimate the disturbance polytope as discussed in Subsection \ref{subsec:distSet_est}, three experiments were carried out with each of the five robots individually. The same control structure as illustrated in Fig. \ref{fig:ctrl_arch_1_with_vars} was employed, but with the trajectory planning module operating in open-loop, i.e., the robot is required to follow a trajectory computed \textit{a priori}.

\begin{figure}[ht!]
	\centering
	\includegraphics[width=0.7\textwidth]{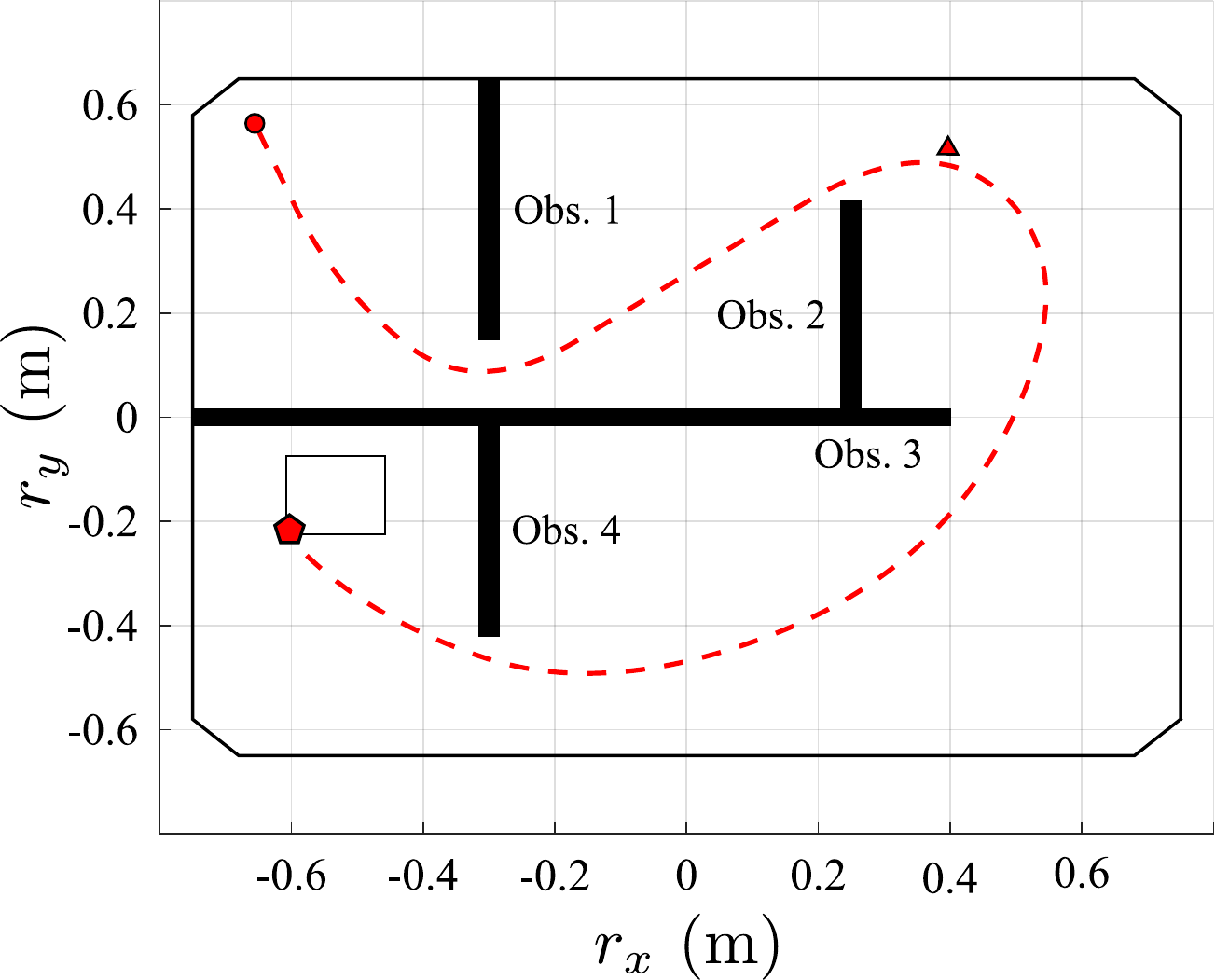}
	\caption{Trajectory and scenario employed in the data collection process. The initial position is represented by a circle. The first phase of the maneuver ends at the triangle marker. The trajectory ends at the pentagon marker.}
	\label{fig:CT_exp_env}     
\end{figure}

The environment depicted in Fig. \ref{fig:CT_exp_env}  was devised such that the robot must reach a target while avoiding four obstacles. In the first phase, the robot is expected to operate in relatively lower velocities due to the placement of obstacles 1 and 2. Thus, noticiable effects of friction are expected. Additionally, the maneuver requires both clockwise and counter-clockwise rotations. In the second phase, which starts from the point labeled with a triangle in Fig. \ref{fig:CT_exp_env}, the robot can develop a higher velocity while avoiding obstacle 4 and reaching the objective.

\begin{figure}[ht!]
	\centering
	\includegraphics[width=1\textwidth]{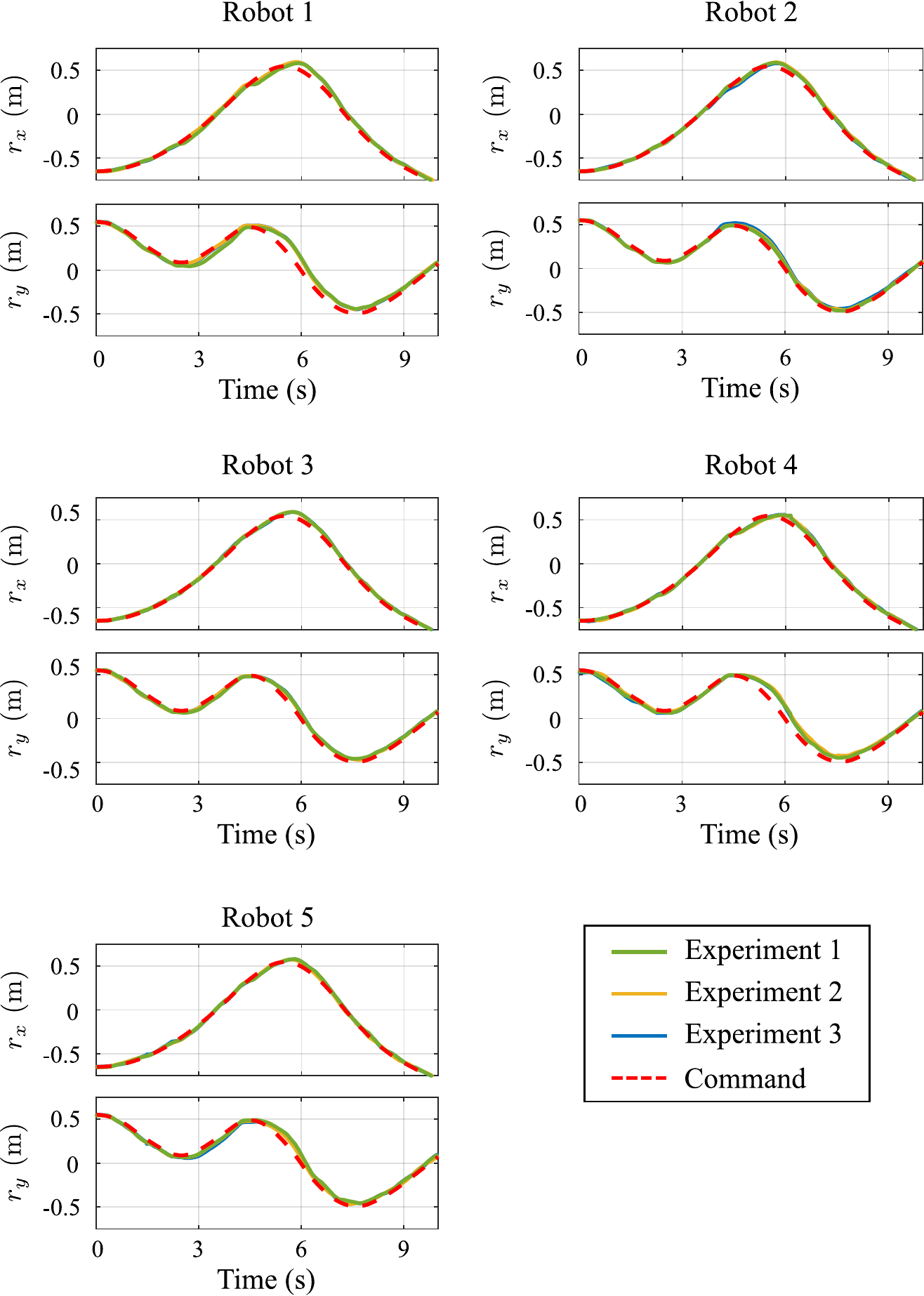}
	\caption{Position tracking results of the data collection process for the robust MPC formulation.}
	\label{fig:r_CT_res}     
\end{figure}

\begin{figure}[ht!]
	\centering
	\includegraphics[width=1\textwidth]{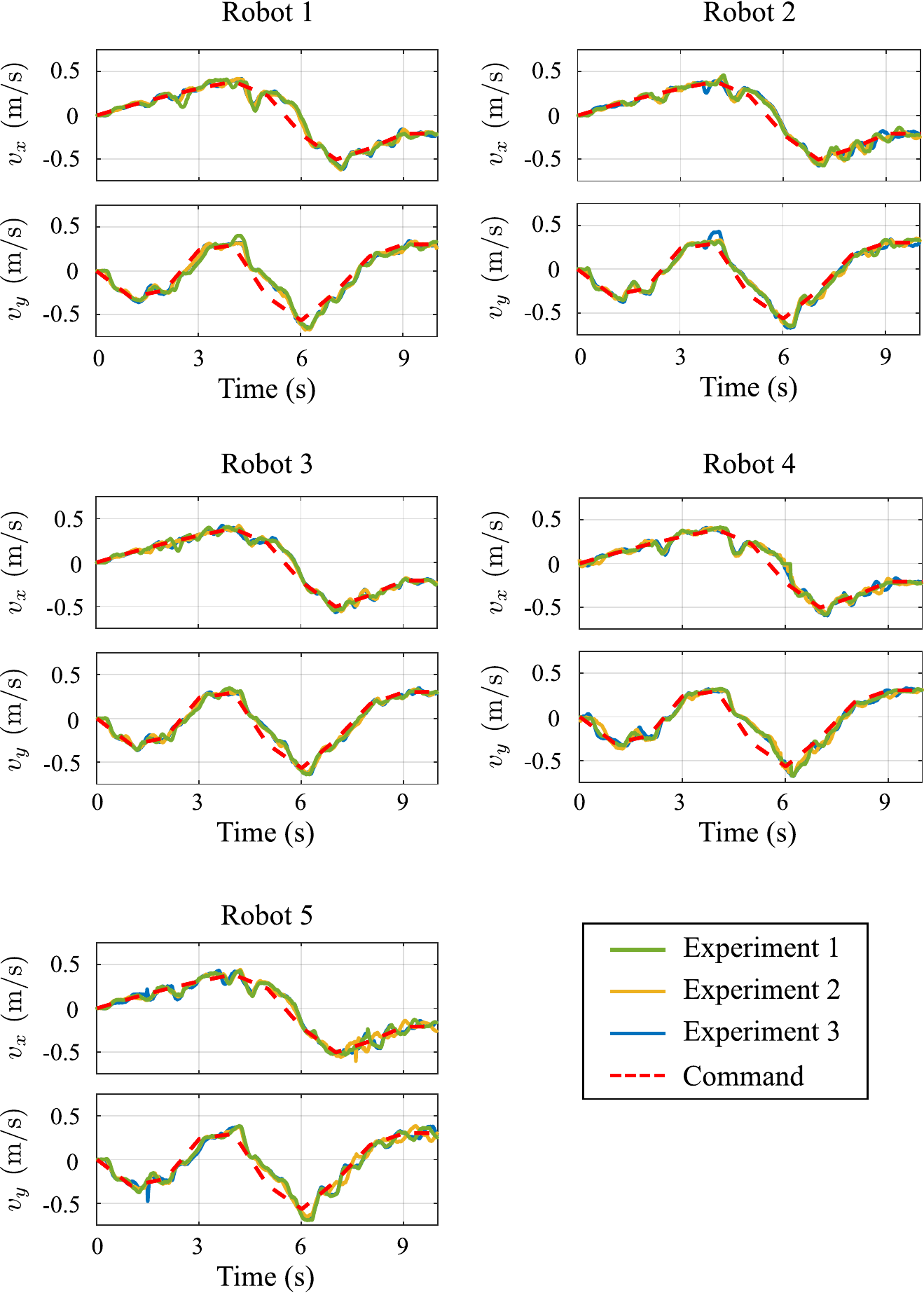}
	\caption{Velocity tracking results of the data collection process for the robust MPC formulation.}
	\label{fig:v_CT_res}     
\end{figure}

The performance of each robot while tracking the trajectory is presented in Figs. \ref{fig:r_CT_res} and \ref{fig:v_CT_res}. The results of the three experiments for each robot were consistent, providing a reliable account of their behavior.
Overall, the proposed controllers performed adequately in terms of position tracking, with small deviations being observed. However, even these slight errors could result in violation of constraints when an optimization-based technique, such as MPC, is employed to compute trajectories without any robustness considerations. 

Adequate, yet worse performance is observed in the velocity tracking results. During the initial time steps, the effect of static friction is clear in the $v_y$ behavior. The ensuing deterioration in the performance is particularly observable until $6$ s when the first phase of the maneuver ends. 

\begin{figure}[ht!]
	\centering
	\includegraphics[width=1\textwidth]{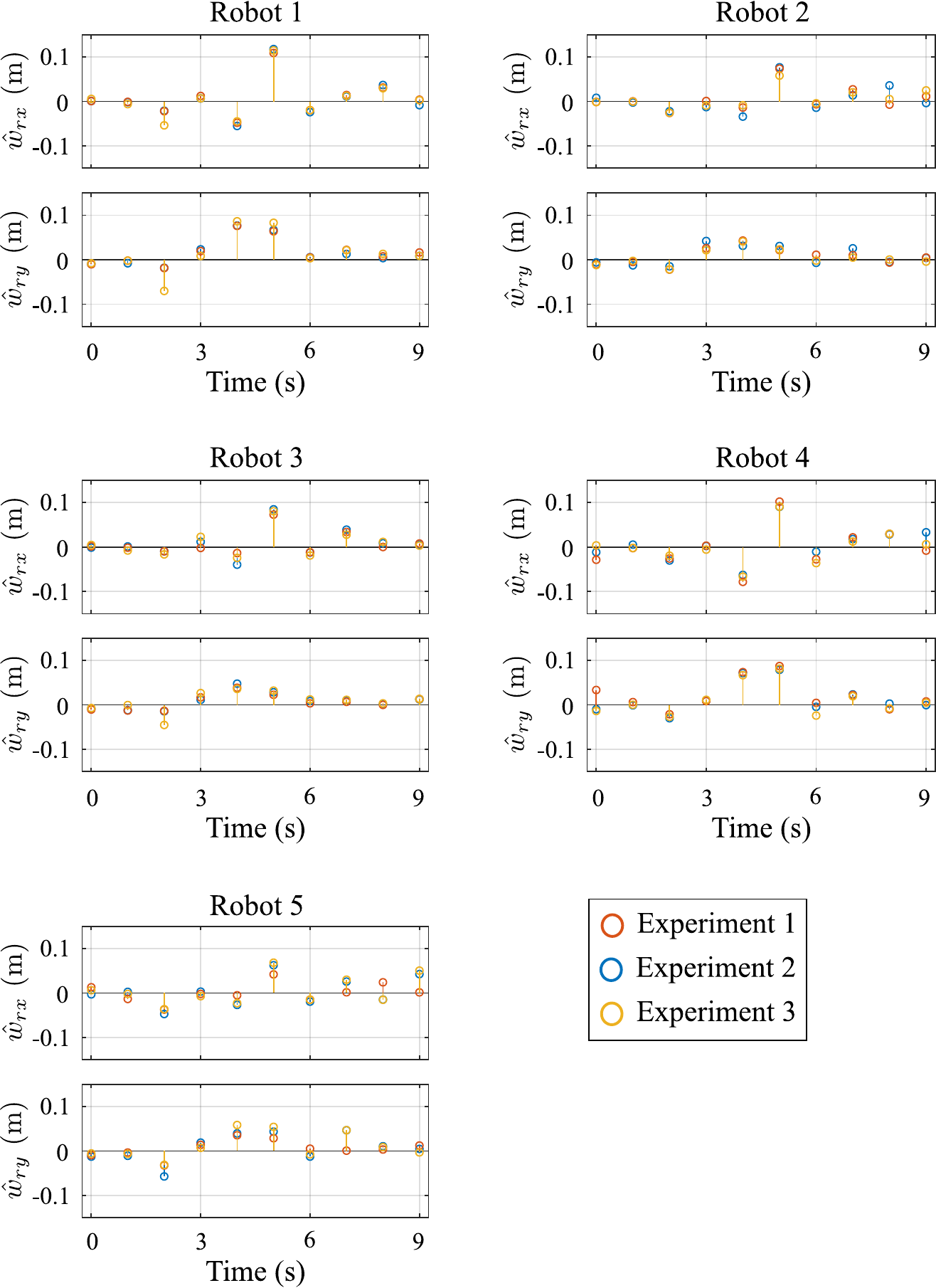}
	\caption{Estimated position disturbances of each agent.}
	\label{fig:wr_res}     
\end{figure}

\begin{figure}[ht!]
	\centering
	\includegraphics[width=1\textwidth]{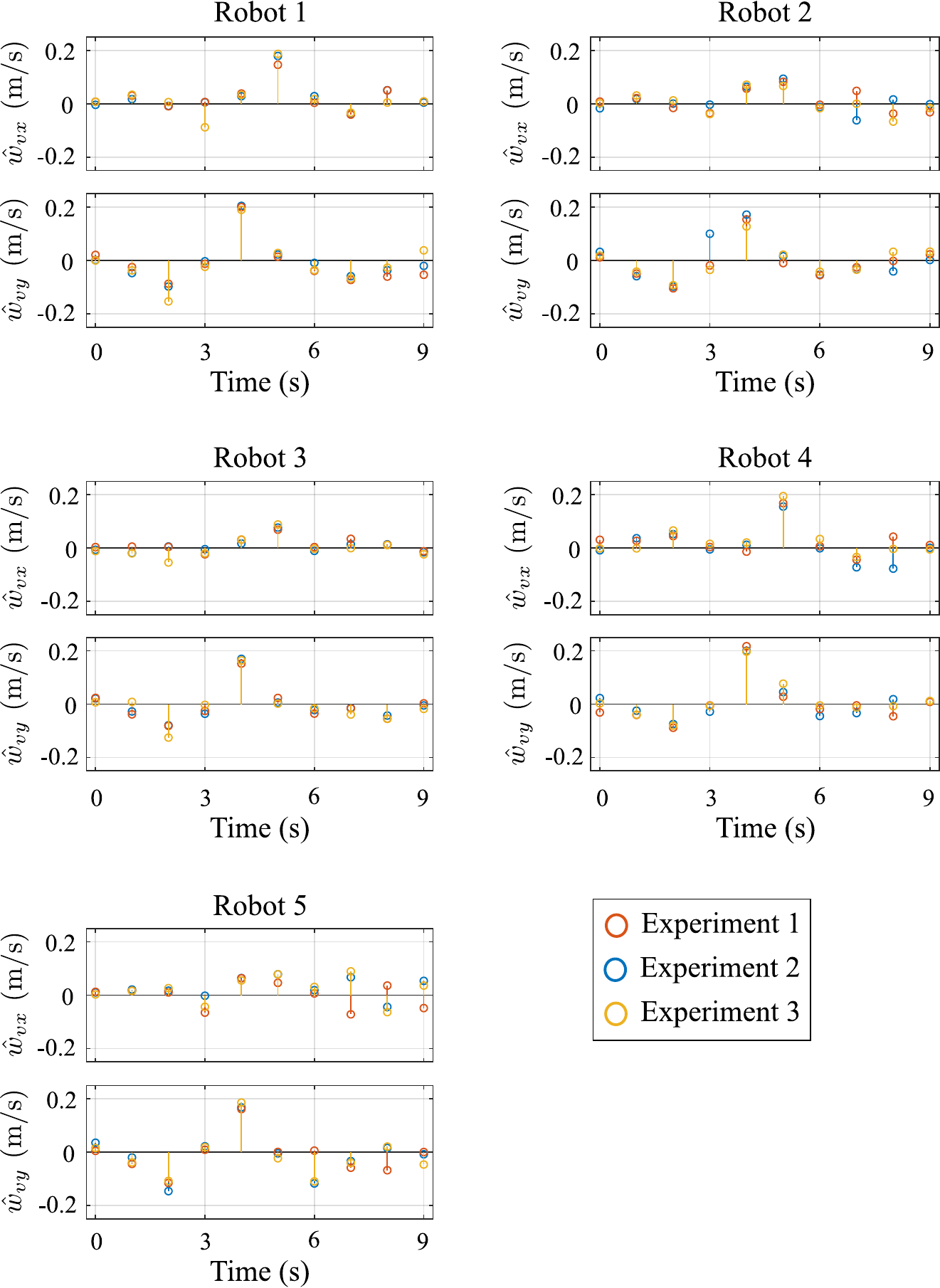}
	\caption{Estimated velocity disturbances of each agent.}
	\label{fig:wv_res}     
\end{figure}

Figures \ref{fig:wr_res} and \ref{fig:wv_res} show the computed disturbance estimate at each time step $k_p$. Generally, the disturbances related to the velocity variables have a higher magnitude. The highest values both for position and velocity disturbances are observed during the time steps related to the first phase of the maneuver, when the robots operated with slower velocities. Thus, these quantities potentially represent the unaccounted friction effects previously reported.

The disturbance polytopes for all robots were computed with the procedure discussed in Subsection \ref{subsec:distSet_est} considering the disturbance estimates of all experiments. The parameters of the resulting polytope $\mathcal{W}_i,\ \forall i \in \mathcal{I}_{n_a}$,  are presented in Table \ref{tab:w_par}.
\begin{table}[!ht]
	\tbl{Estimated disturbance polytope parameters for the five robots.}{
		\begin{tabular}{ccccc}
			\bottomrule
			Robot & $\sigma_{rx}$ (m) & $\sigma_{vx}$ (m/s) & $\sigma_{ry}$ (m) & $\sigma_{vy}$ (m/s) \\ \midrule
			1     & 0.1178        & 0.1869        & 0.0865        & 0.2047        \\ 
			2     & 0.0769        & 0.0937        & 0.0431        & 0.1719        \\ 
			3     & 0.0846        & 0.0888        & 0.0476        & 0.1697        \\ 
			4     & 0.1020        & 0.1952        & 0.0872        & 0.2171        \\ 
			5     & 0.0680        & 0.0894        & 0.0584        & 0.1860        \\ \bottomrule
		\end{tabular}}
	\label{tab:w_par}
\end{table}


\section{Experiments} \label{sec:expRes}

\subsection{Setup}

\begin{figure}[ht!]
	\centering
	\includegraphics[width=0.5\textwidth]{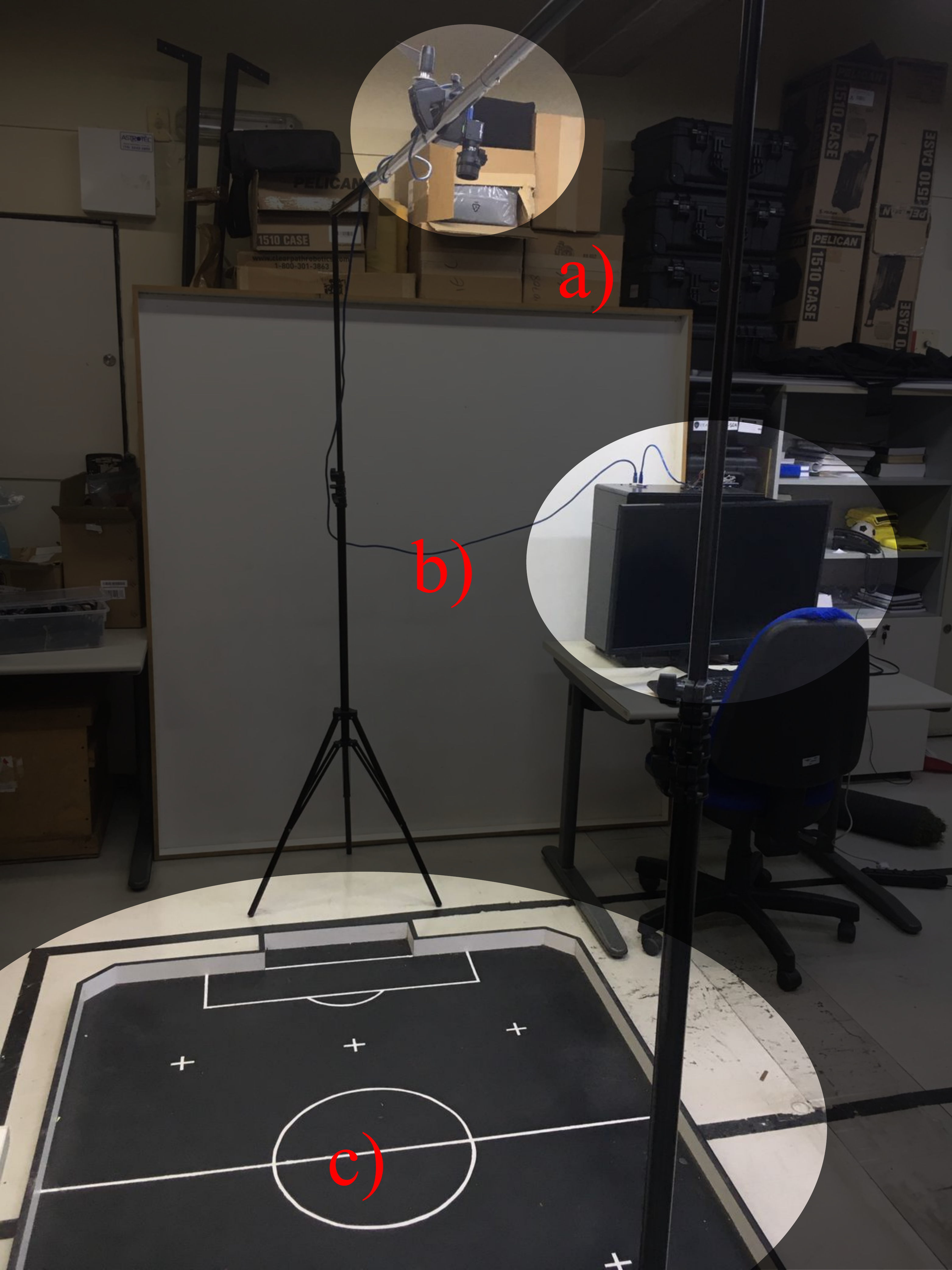}
	\caption{a) Camera used for the position measurement. b) Machine employed for computation of the algorithms. c) Soccer field where the robots operate.}
	\label{fig:exp_setup_photo}     
\end{figure}

The experimental demonstrations in this work were carried out in the Autonomous Computational Systems Laboratory (LAB-SCA) at ITA. The MAS was comprised of the same five robots used in Section \ref{section:robust_mpc}.
 A picture of the experimental setup is presented in Fig. \ref{fig:exp_setup_photo}. Three main components can be observed: the camera in Fig. \ref{fig:exp_setup_photo}a,  the computer in Fig. \ref{fig:exp_setup_photo}b, and the soccer field in Fig. \ref{fig:exp_setup_photo}c. 

 The camera employed for the measurement of the agents' positions and orientations is a Point Grey$^{\text{\textregistered}}$ Flea3 operating with a frequency of 60 Hz due to the high computational cost of the computer vision algorithms.

 Only virtual obstacles were considered during the experiments, i.e., no physical obstacles were placed in the field for the sake of simplicity. This choice does not influence the proposed algorithms in any way and the same results would be observed in an environment with equivalent physical obstacles.
 The same initial conditions and parameters as in the simulation presented in Subsection \ref{subsec:simVSS} were employed in the experiments.  The main computer was equipped with an Intel$^{\text{\textregistered}}$ Core i9-7900X 3.30 GHz CPU and 64 GB of RAM.  Video footage of the experiments is available online\footnote{Scenario 1: \url{https://youtu.be/fyb47v_3n_s?t=15} ; Scenario 2:  \url{https://youtu.be/fyb47v_3n_s?t=34} }. 

 \subsection{Scenarios}
\begin{figure}[ht!]
 	\centering
 	\includegraphics[width=1\textwidth]{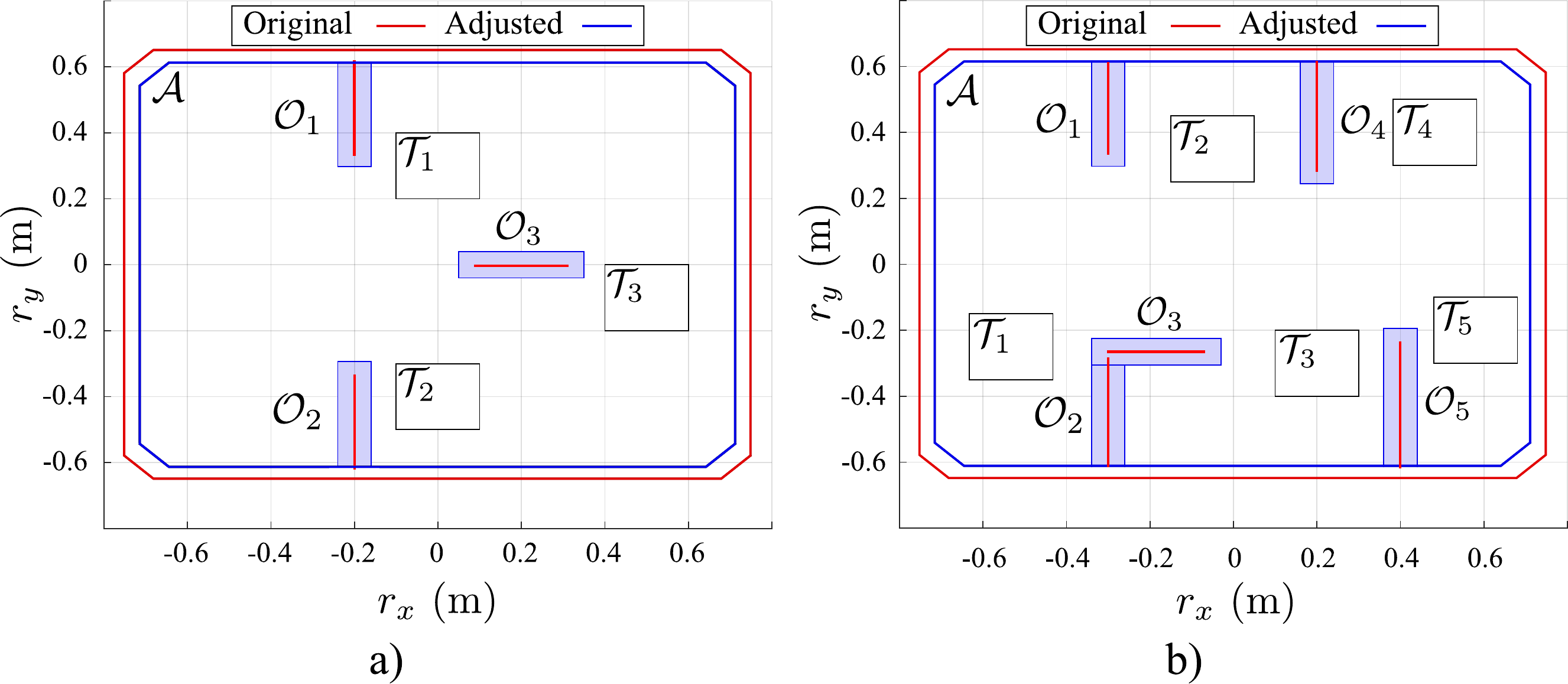}
 	\caption{a)  Scenario 1, b) scenario 2 and their corresponding versions adjusted to account for the body of the robots. Obstacles and targets are represented by the sets $\mathcal{O}_g,\ \forall g \in \mathcal{I}_5$, $\mathcal{T}_e,\ \forall e \in \mathcal{I}_5$, respectively.}
 	\label{fig:scen1}     
 \end{figure}

 The first scenario is comprised of $n_o=3$ obstacles and $n_t=3$ targets, as illustrated by Fig. \ref{fig:scen1}a. Since the constraints are applied to the geometric center of the robots, it is necessary to reduce the size of the field and increase the size of the obstacles considering the dimensions of the robots to prevent collisions. The  ensuing adjusted environment is also depicted in Fig. \ref{fig:scen1}a.

 Scenario 1 was devised to illustrate the capacity of the proposed system to effectively coordinate the MAS in a constrained space and solve the task allocation problem, where targets 1 and 2 are optional, whereas target 3 is mandatory. Obstacles 1 and 2 create a narrow corridor requiring appropriate coordination to be overcome by the group without collisions; obstacle 3 provides an additional challenge to the mission completion as it blocks the path between the robots and the terminal target.
 
  A more convoluted environment is proposed in scenario 2, as presented in Fig. \ref{fig:scen1}b. There are five obstacles and targets distributed throughout the field, resulting in a more challenging mission in terms of the task allocation and coordination of agents.

\subsection{Results of Scenario 1}

\begin{figure}[ht!]
	\centering
	\includegraphics[width=0.9\textwidth]{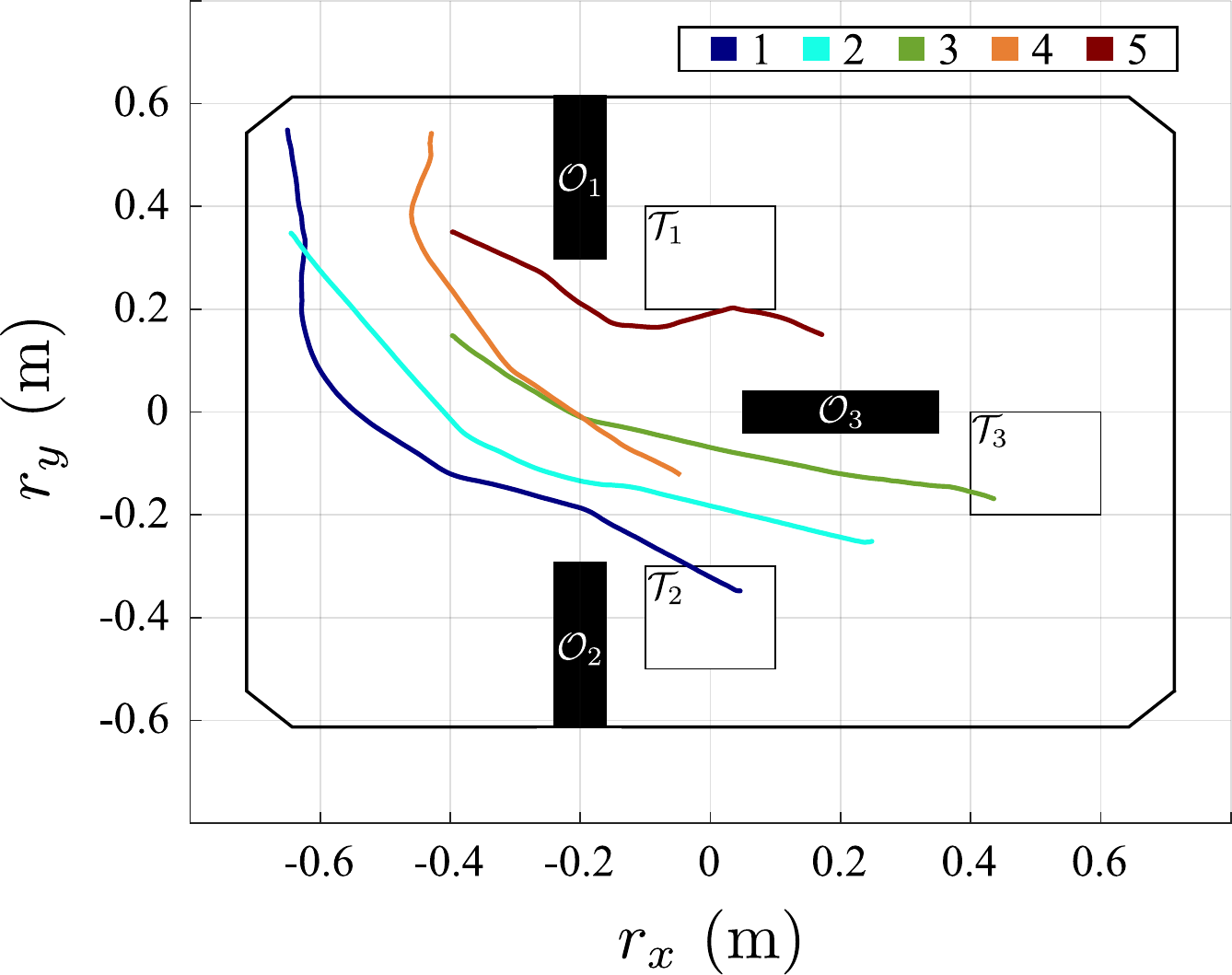}
	\caption{Results of the experiments in scenario 1 using the robust MPC formulation.}
	\label{fig:traj_exp1}     
\end{figure}

\begin{figure}[ht!]
	\centering
	\includegraphics[width=1\textwidth]{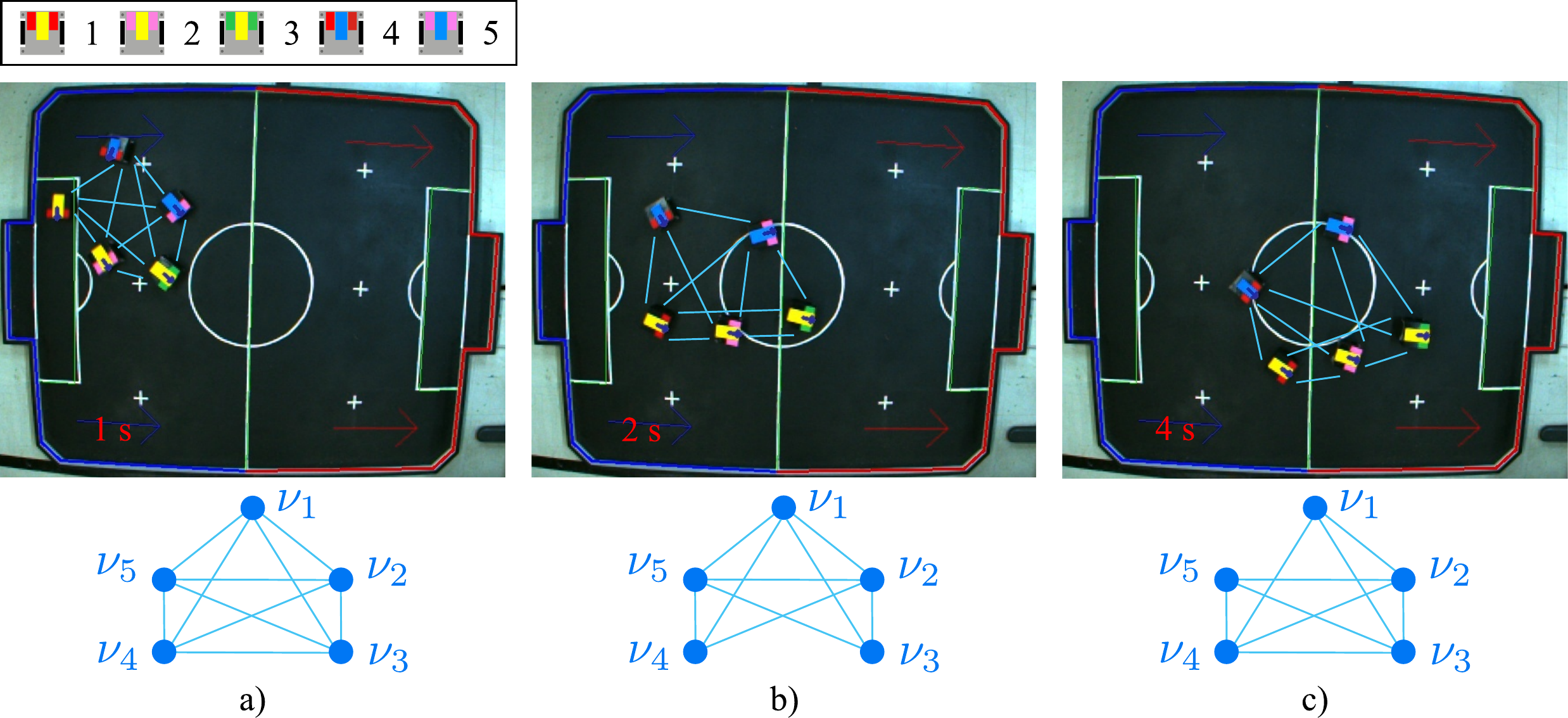}
	\caption{ Robots during experiment in scenario 1 at a) $t=1$ s, b) $t=2$ s, and c) $t=4$ s and corresponding graphs representing the communication network of the MAS.}
	\label{fig:traj_frames_exp1}     
\end{figure}

The trajectories followed by each agent in the experiment related to scenario 1 are presented in Fig. \ref{fig:traj_exp1}. The robots were able to coordinate to finish the mission without collisions while visiting two extra targets and maintaining the robust connectivity of the communication network. Fig. \ref{fig:traj_frames_exp1}a shows the configuration of the robots after 1 s; they remain cluttered as the group carefully maneuvers to surpass the narrow corridor created by obstacles 1 and 2, resulting in a communication network represented by a complete graph. After 2 s, the third robot develops a lead towards the terminal target $\mathcal{T}_3$, disconnecting from robot 4, as shown in Fig. \ref{fig:traj_frames_exp1}b. Notice that robots 1 and 3 conserve the connection between them due to the octagonal geometry of the connectivity region. At the terminal time step, illustrated by Fig. \ref{fig:traj_frames_exp1}c, the MAS was able to spread, visiting all targets and maintaining the robust connectivity of the communication network. The maneuver finished after 4 seconds and the MAS remained robustly connected during the entire mission. 

\begin{figure}[ht!]
	\centering
	\includegraphics[width=1\textwidth]{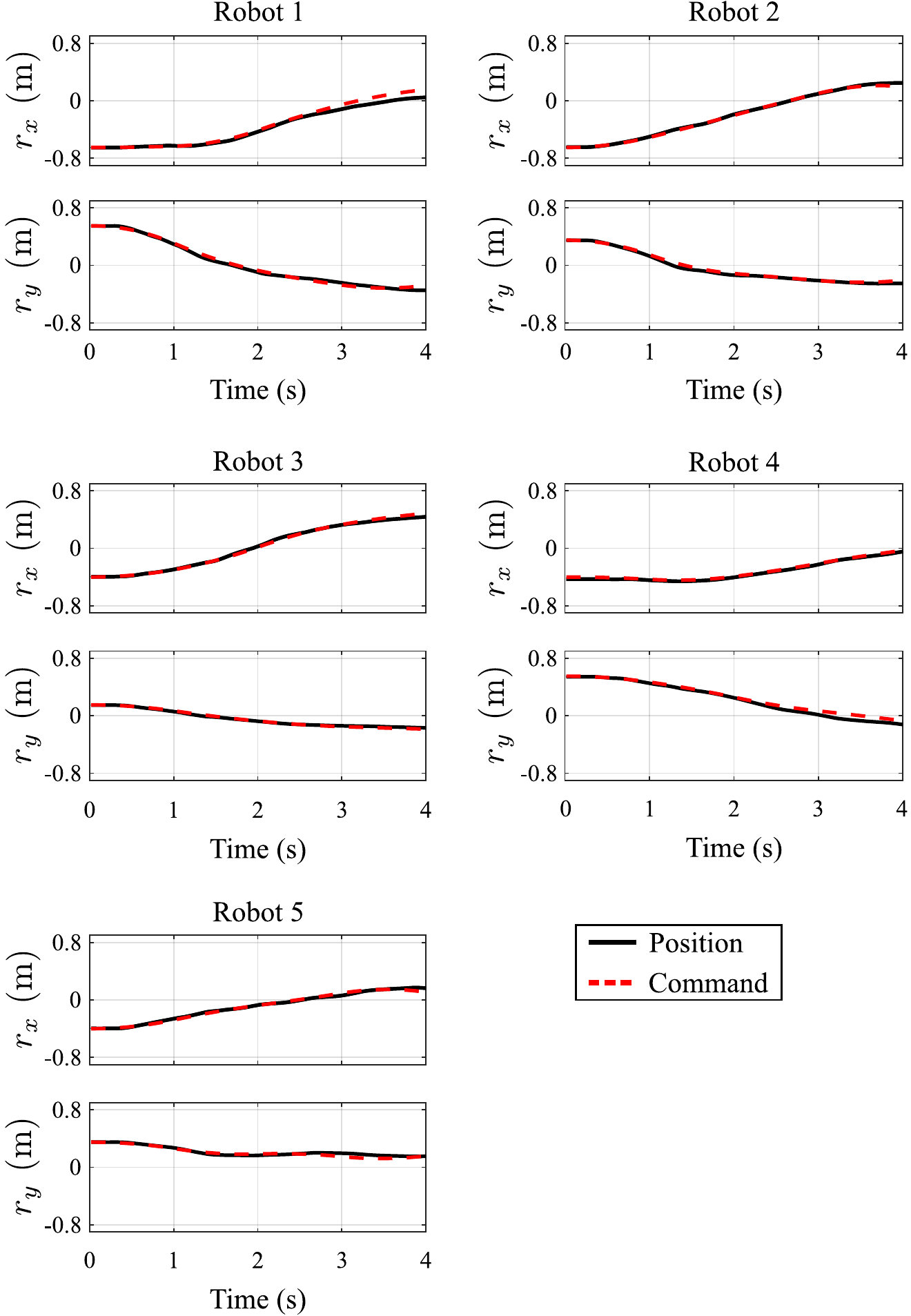}
	\caption{ Position tracking results for each agent during the experiment in scenario 1.}
	\label{fig:pos_exp1}      
\end{figure}

\begin{figure}[ht!]
	\centering
	\includegraphics[width=1\textwidth]{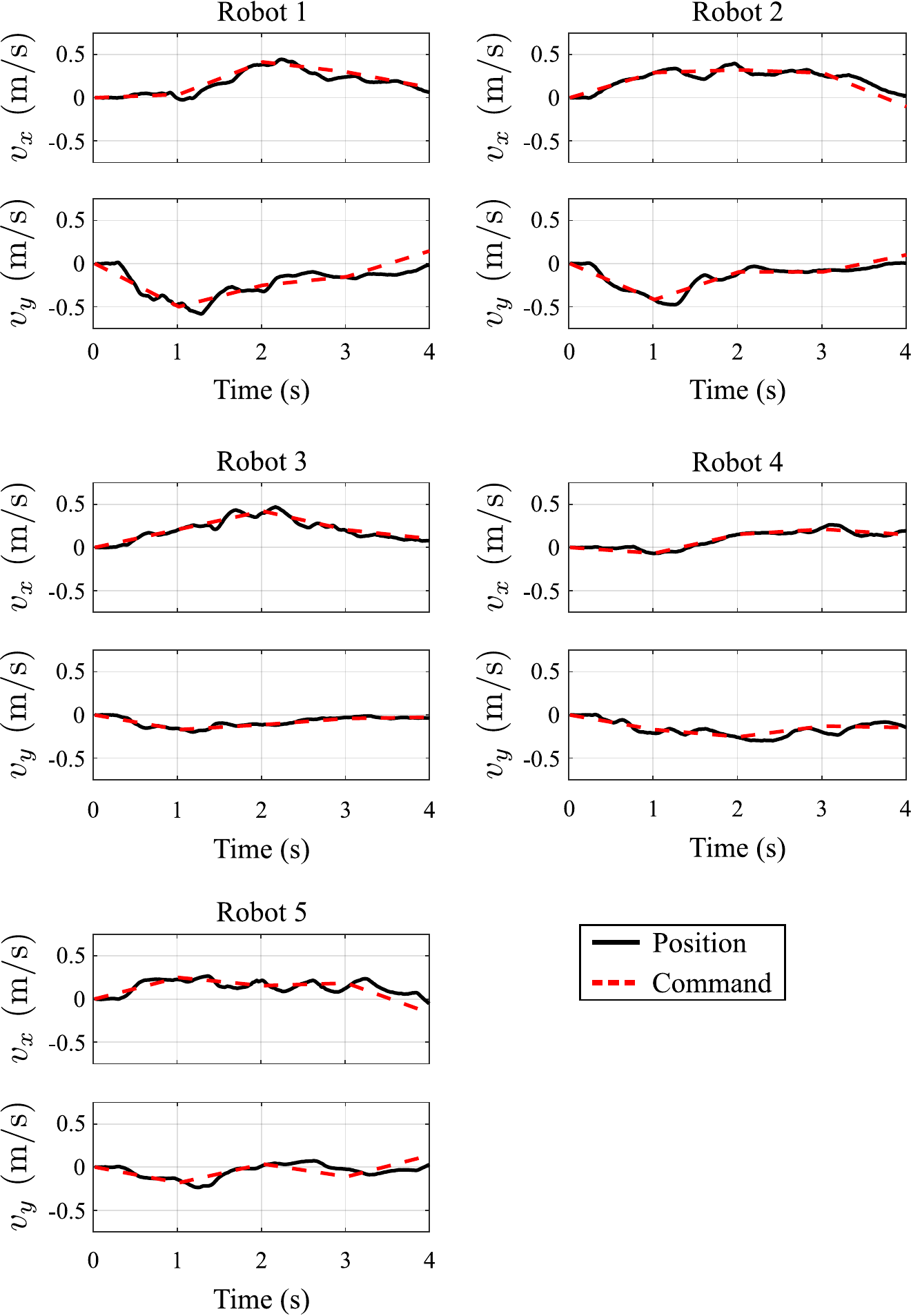}
	\caption{Velocity tracking results for each agent during the experiment in scenario 1.}
	\label{fig:vel_exp1}      
\end{figure}

\begin{figure}[ht!]
	\centering
	\includegraphics[width=1\textwidth]{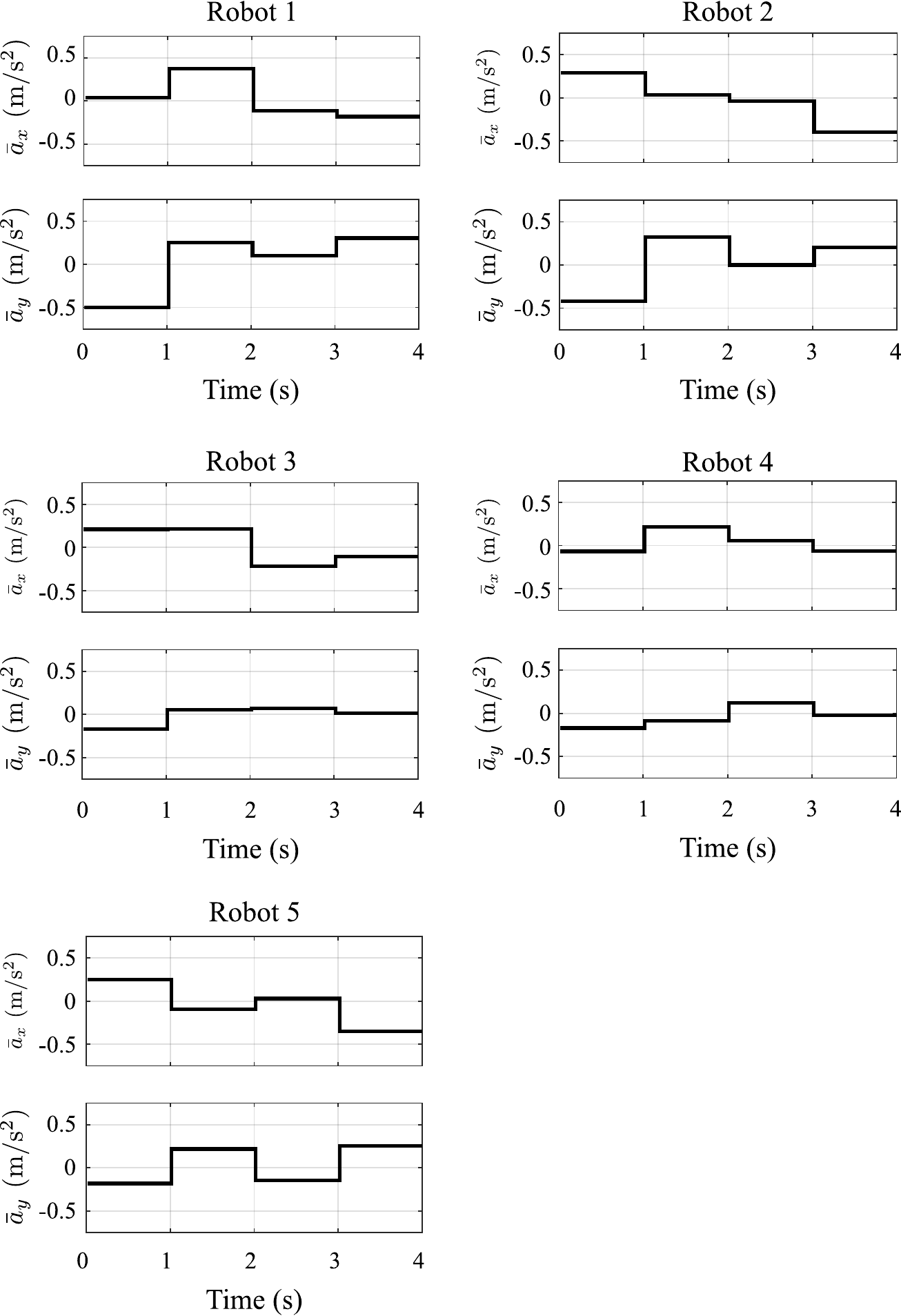}
	\caption{Acceleration commands for each agent during the experiment in scenario 1.}
	\label{fig:acc_exp1}      
\end{figure}

 The positions and velocities developed by each robot as well as the corresponding interpolated commands are depicted in Figs. \ref{fig:pos_exp1} and \ref{fig:vel_exp1}. The acceleration commands provided by the MPC-MIP planner are presented in Fig. \ref{fig:acc_exp1}. The proposed controllers provided adequate position tracking performance for all robots. The velocity results show slightly worse results mainly due to friction. The effects of static friction can be particularly observed in the results of $v_y$ for the first robot. It remained unresponsive to the initial commands for about $0.3$ s until enough error was accumulated yielding control signals with enough magnitude to overcome the friction. The sudden burst resulted in an overshoot at approximately $0.5$ s. 
 In spite of the observed tracking inaccuracies, the system was able to operate appropriately mainly due to the margins added by the robust MPC formulation.


\subsection{Results of Scenario 2}
\begin{figure}[ht!]
	\centering
	\includegraphics[width=0.9\textwidth]{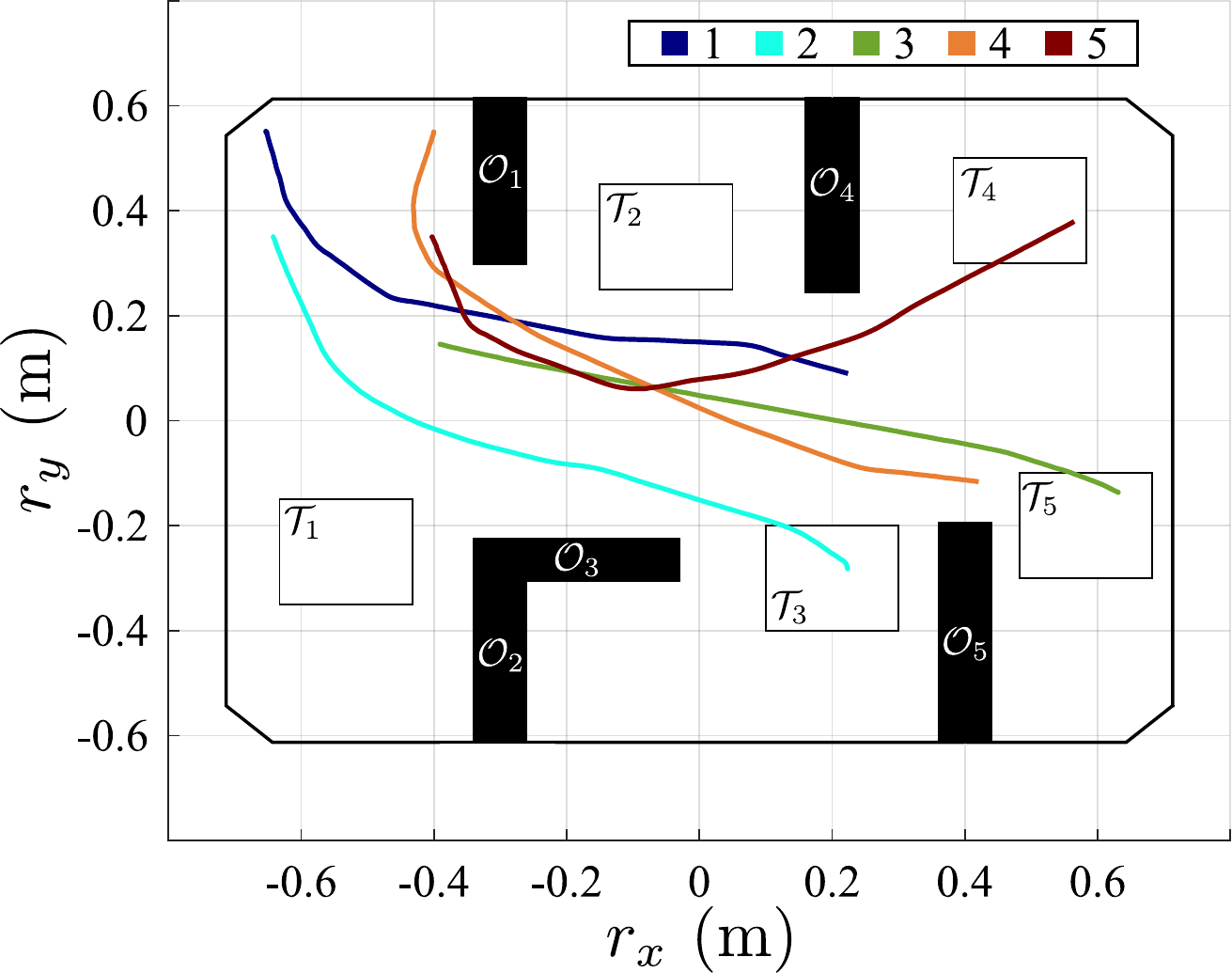}
	\caption{Results of the experiments in scenario 2 using the robust MPC formulation.}
	\label{fig:traj_exp2}     
\end{figure}

The trajectories followed during the experiment in scenario 2 are shown in Fig. \ref{fig:traj_exp2}. This environment was more challenging in terms of task allocation due to the number and distribution of the targets. The MAS was successful in reaching the terminal target while preventing collisions and conserving the robust connectivity of its communication network. The decision process is illustrated by the choice of only collecting the rewards associated with targets 3 and 4. The maneuver finished after 4 seconds.

\begin{figure}[ht!]
	\centering
	\includegraphics[width=1\textwidth]{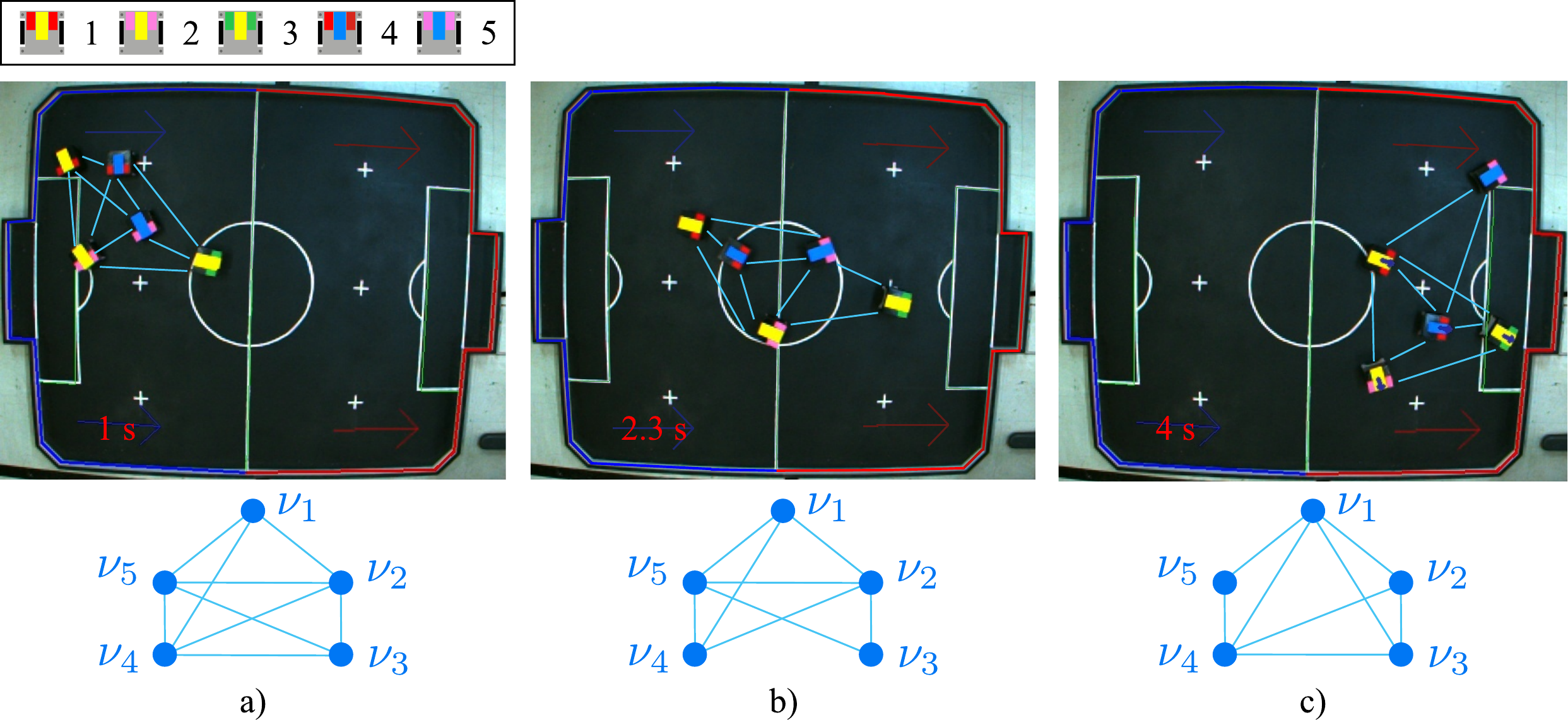}
	\caption{ Robots during experiment in scenario 2 at a) $t=1$ s, b) $t=2.3$ s, and c) $t=4$ s and corresponding graphs representing the communication network of the MAS.}
	\label{fig:traj_frames_exp2}     
\end{figure}

The configuration of the robots during the experiment at $t=1$ second is presented in Fig. \ref{fig:traj_frames_exp2}a. The third robot moves towards the last target disconnecting itself from the first robot on the upper left side. The next frame depicts the state of the MAS at $t=2.3$ s is shown in Fig. \ref{fig:traj_frames_exp2}b. Robot 3 remains moving towards $\mathcal{T}_5$ while robots 1 and 4 coordinate to surpass the narrow corridor, resulting in the disconnection of robot 3 from robots 1 and 4. Nevertheless, the corresponding communication network remains robustly connected. Finally, the last frame representing the system at $t=4$ s is presented in Fig. \ref{fig:traj_frames_exp2}c. Robots 2 and 5 reached targets 3 and 4, respectively; the mandatory target was reached by robot 3. The MAS was stretched considerably, as evidenced by the fact that robot 5 remains connected to robot 4 but is disconnected from robot 3 even though robots 3 and 4 are only slightly distanced vertically. This implies that agents 5 and 4 are positioned at the boundaries of each other's communication regions.

\begin{figure}[ht!]
	\centering
	\includegraphics[width=1.\textwidth]{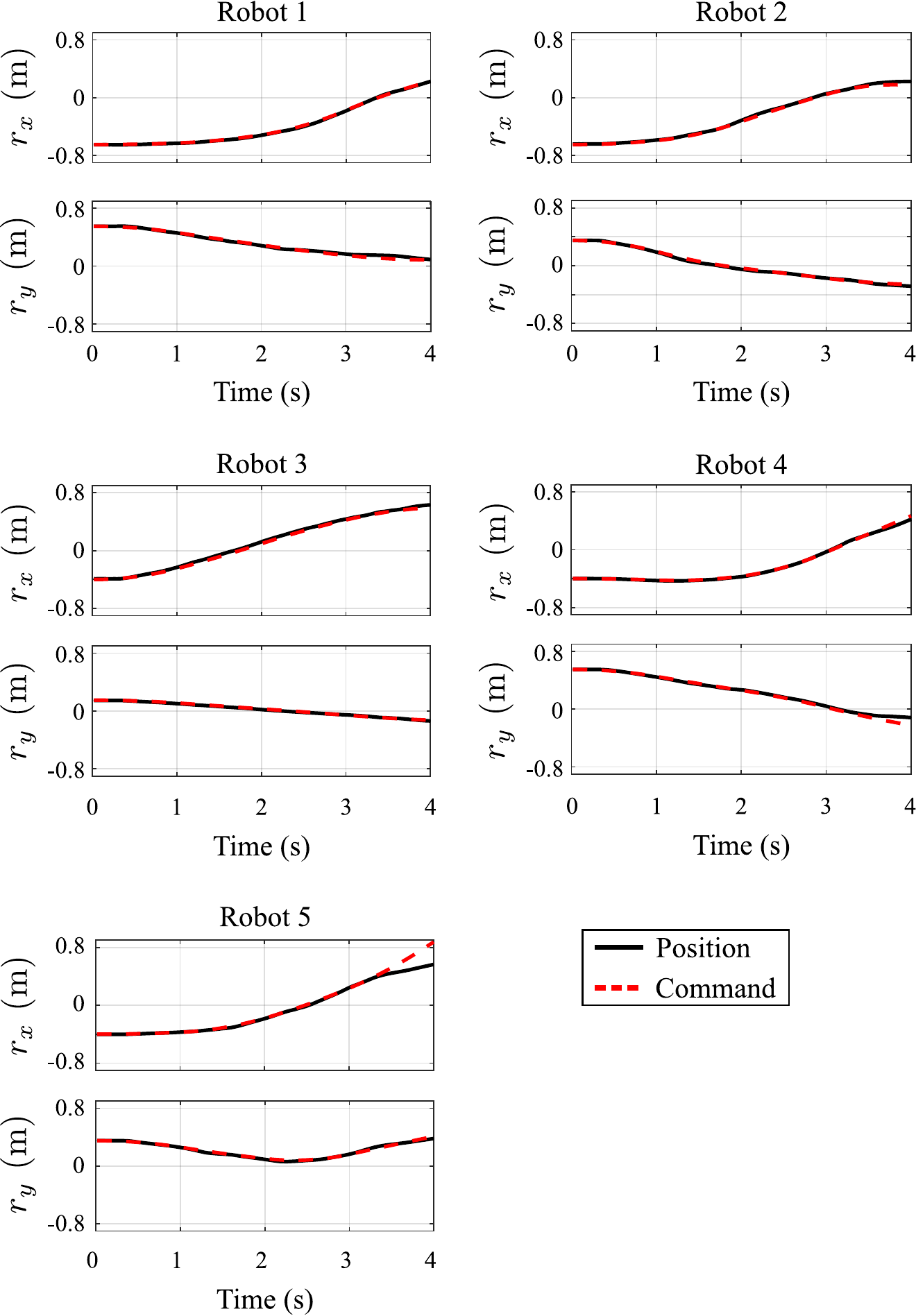}
	\caption{Position tracking results for each agent during the experiment in scenario 2.}
	\label{fig:pos_exp2}     
\end{figure}

\begin{figure}[ht!]
	\centering
	\includegraphics[width=1.\textwidth]{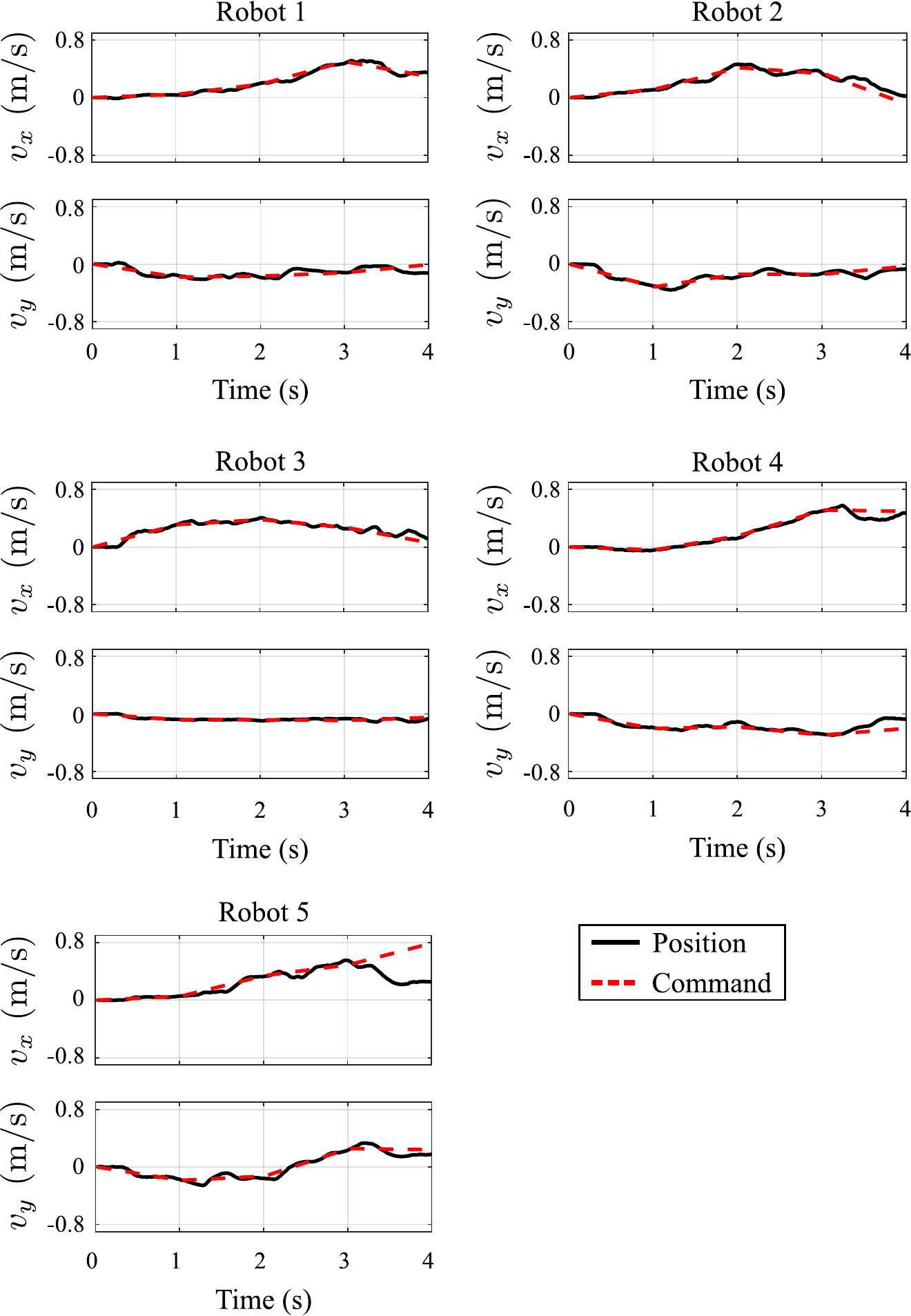}
	\caption{Velocity tracking results for each agent during the experiment in scenario 2.}
	\label{fig:vel_exp2}     
\end{figure}

\begin{figure}[ht!]
	\centering
	\includegraphics[width=1.\textwidth]{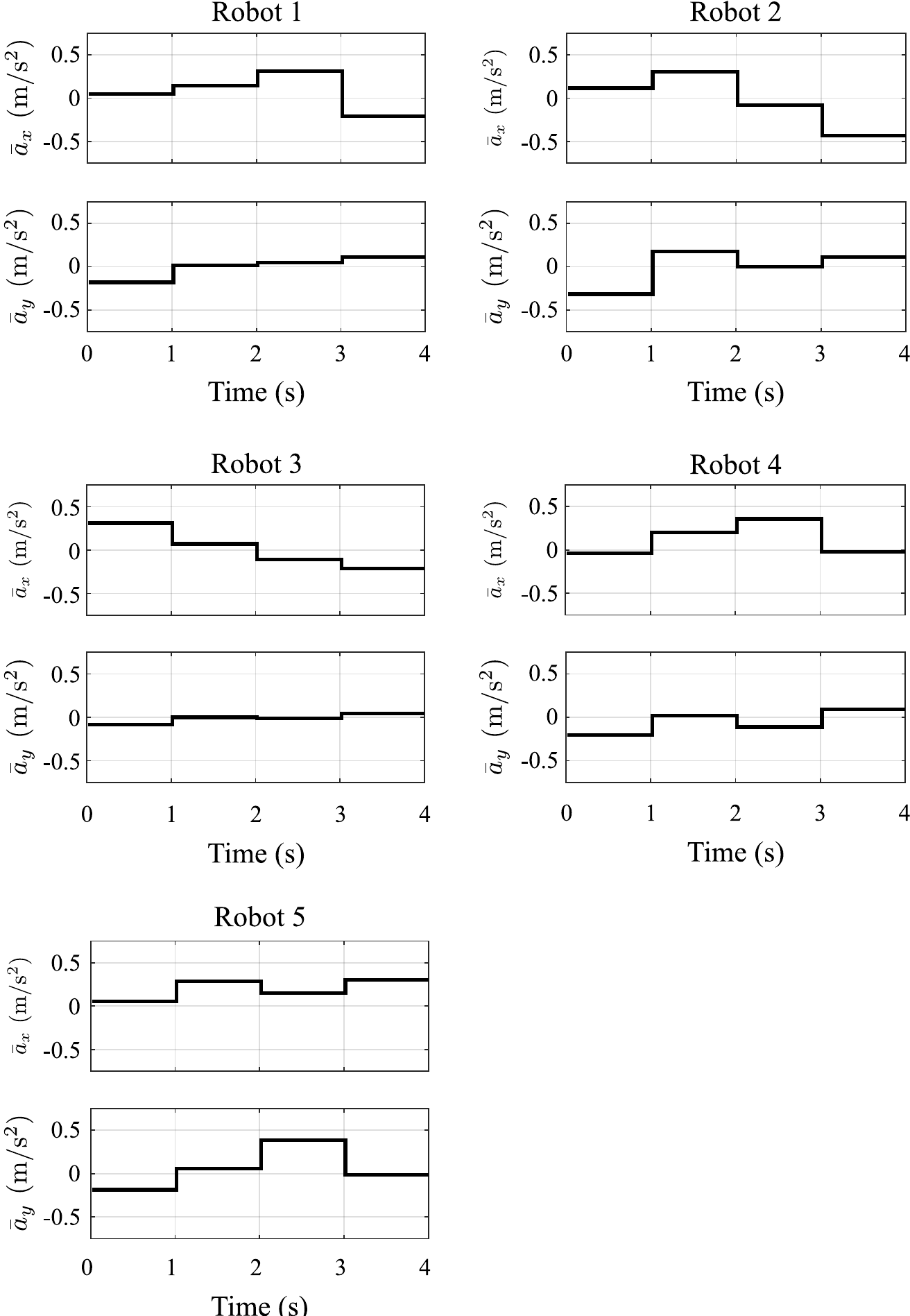}
	\caption{Acceleration commands for each agent during the experiment in the scenario 2.}
	\label{fig:acc_exp2}     
\end{figure}

The position and velocity of each robot are presented in Figs. \ref{fig:pos_exp2} and \ref{fig:vel_exp2}, respectively. The acceleration commands  are shown in Fig. \ref{fig:acc_exp2}. The system was able to track the position commands satisfactorily. The velocity tracking performance was slightly better in scenario 2 compared to scenario 1, as the robots generally developed higher velocities rendering less pronounced friction effects. Robots 4 and 5 present a considerable disturbance to their velocity at about $t=3$ s which is potentially associated with irregularities in the field that result in areas with higher friction. Nevertheless, the MAS was able to complete its mission with no collisions and conserve the robustness of its communication network.


\section{Conclusion}\label{sec:conclusion}

In this work, an MPC-based motion planner was employed as part of a hierarchical control structure used to steer a group of differential robots towards mission completion under connectivity constraints. A robust MPC-MIP formulation provided further robustness to the trajectories planned. A data-based scheme was proposed to estimate the disturbances sets required by this formulation. The experiments have shown, in two distinct obstacle-filled scenarios, that the control architecture was successful in safely steering, guiding, and allocating tasks in real-time for the MAS. Future work could be focused on improving the disturbance set estimation method to reduce its conservatism and experimental evaluations of the scheme under line of sight connectivity requirements.

\section*{Acknowledgments}
This study was financed in part by the Coordena\c c\~ao de Aperfei\c coamento de Pessoal de N\' ivel Superior - Brasil (CAPES) - Finance Code 001.

\theendnotes

\bibliographystyle{tfnlm}
\bibliography{interactnlmsample}

\end{document}